\documentclass[11pt, a4paper, onecolumn, copyright]{AweAI}

\usepackage{multirow}
\usepackage[authoryear, round]{natbib} 
\usepackage{graphicx}
\usepackage{microtype}
\usepackage{subcaption}
\usepackage{booktabs}
\usepackage{listings} 
\usepackage{makecell}

\usepackage[most]{tcolorbox}
\tcbuselibrary{listings}
\usepackage{xspace}
\usepackage{textcomp}
\usepackage{algorithm}
\usepackage{algpseudocode}
\usepackage{amsmath}
\usepackage{amssymb}
\usepackage{mathtools}
\usepackage{amsthm}
\usepackage{enumitem}
\usepackage{pifont}
\definecolor{darkgreen}{RGB}{50,100,0}
\definecolor{darkred}{RGB}{200, 0, 0}

\usepackage{xurl}
\usepackage{titletoc}
\usepackage{needspace}

\usepackage{float}       
\usepackage{placeins}    
\usepackage{tabularx}    
\usepackage{array}       

\definecolor{tableblue}{HTML}{E6EFFA}
\definecolor{calibrationorange}{HTML}{FFF0E3}
\definecolor{headergray}{HTML}{444444}

\colorlet{DarkGreen}{green!50!black}
\colorlet{DarkRed}{red}

\DeclareRobustCommand{\github}{%
  \raisebox{-1.5pt}{%
    \includegraphics[height=1.05em]{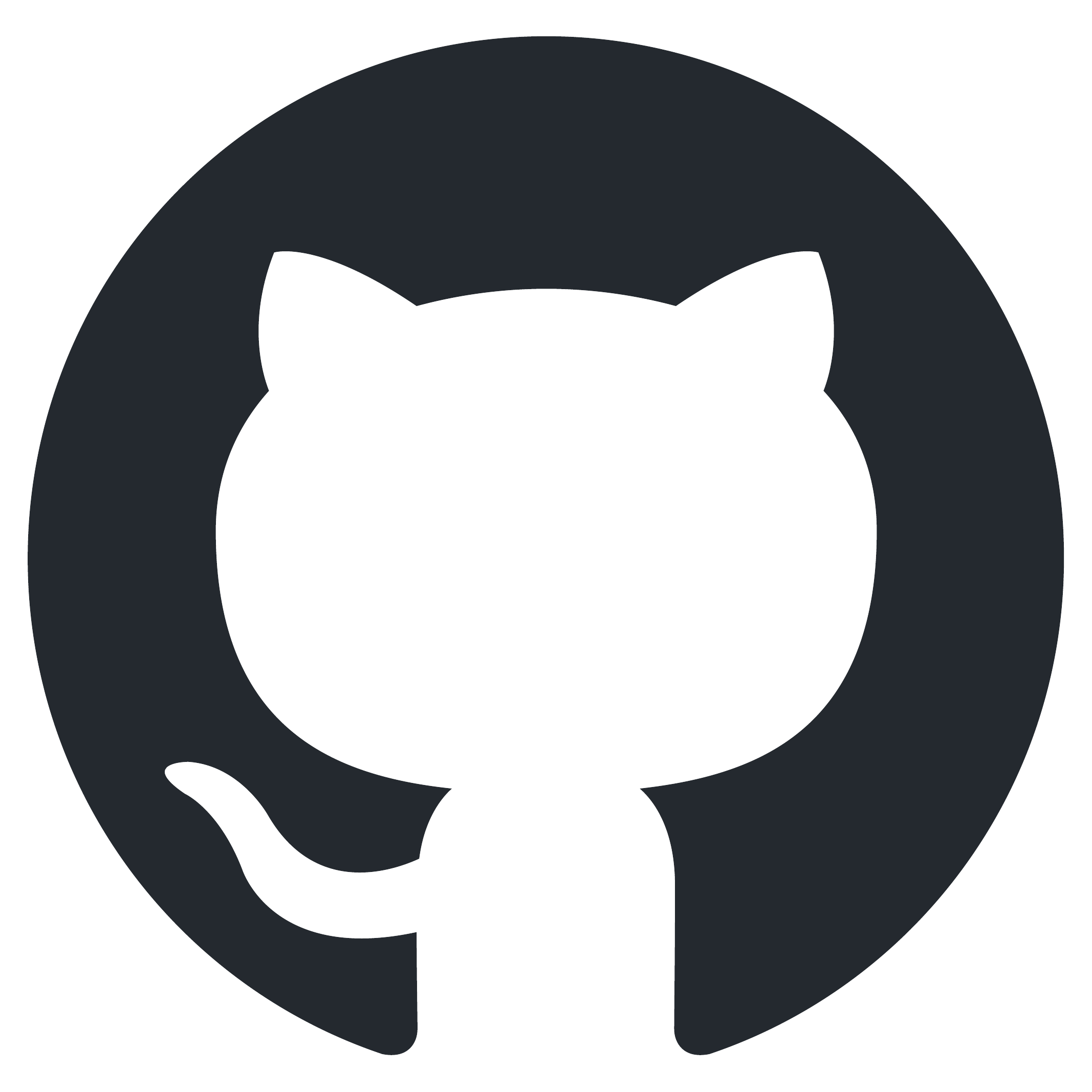}%
  }\xspace
}

\DeclareRobustCommand{\huggingface}{%
  \raisebox{-1.5pt}{%
    \includegraphics[height=1.05em]{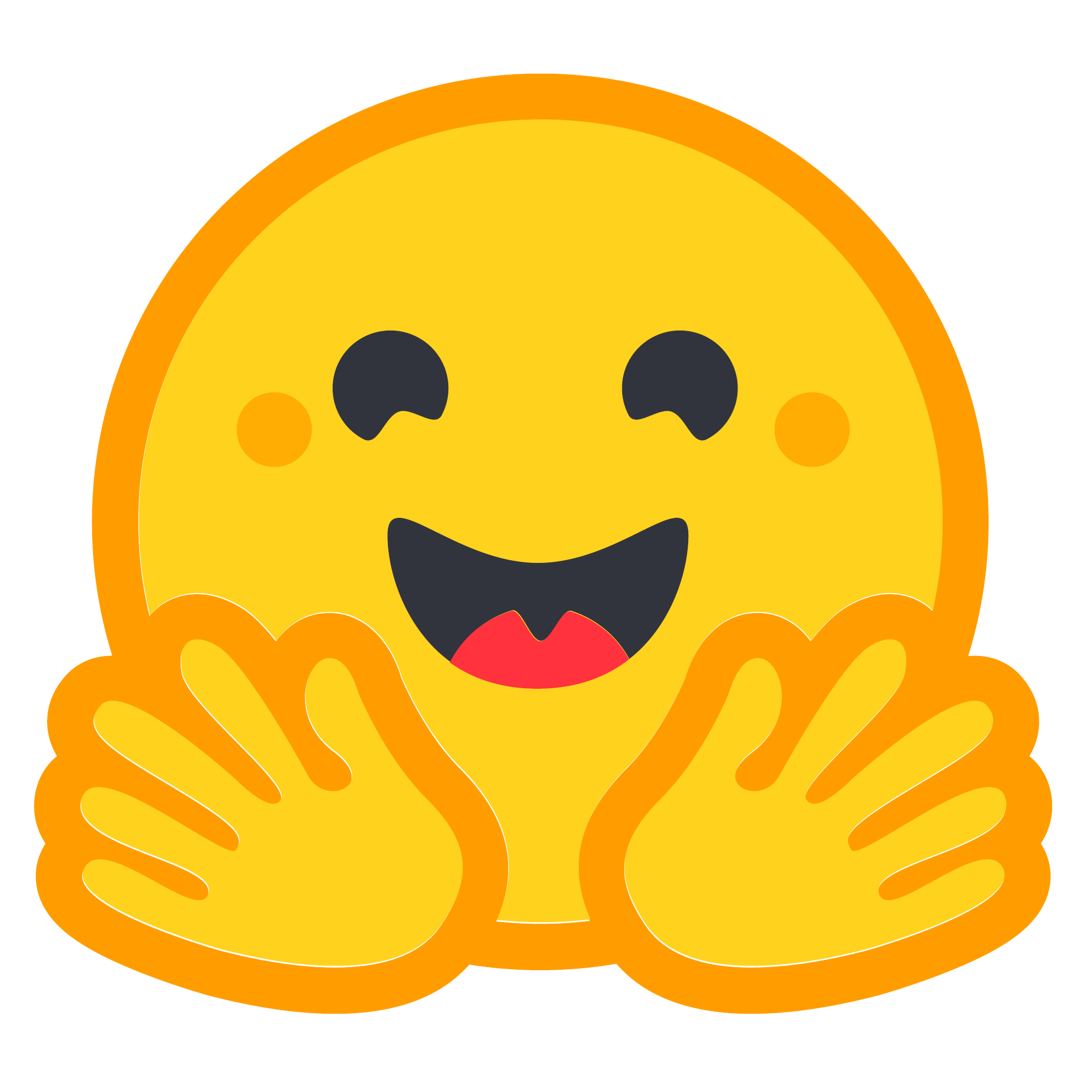}%
  }\xspace
}

\newcommand{\resourcebuttons}{%
  \huggingface\
  \href{https://huggingface.co/datasets/AweAI-Team/CalibForge}{Dataset}%
  \hspace{1.5em}%
  \github\
  \href{https://github.com/AweAI-Team/CalibForge}{GitHub}%
}

\newcommand{\papername}{CalibForge}
\newcommand{\scaffold}{CalibForge-Eval}
\newcommand{\msem}[2]{#1\thinspace\ensuremath{\pm}\thinspace#2}
\newcommand{\algphase}[2]{%
  \Statex \colorbox{#1}{\strut\textbf{#2}}}

\definecolor{caseblue}{HTML}{3F6288}
\definecolor{caseblueback}{HTML}{F2F7FF}
\definecolor{casegreen}{HTML}{397A57}
\definecolor{casegreenback}{HTML}{F2F8F4}
\definecolor{casered}{HTML}{8F3F3F}
\definecolor{casegray}{HTML}{6F7477}
\definecolor{casecontrast}{HTML}{C46A6A}
\definecolor{casemulti}{HTML}{5D7FA6}
\definecolor{caseorange}{HTML}{D7A23A}
\definecolor{caseorangeback}{HTML}{FFF8E8}
\definecolor{casegrayback}{HTML}{F4F5F5}

\newtcolorbox{takeawaybox}{
  enhanced,
  colback=caseblueback,
  colframe=caseblueback,
  boxrule=0pt,
  arc=1.5mm,
  left=2.5mm,
  right=2.5mm,
  top=1.5mm,
  bottom=1.5mm,
  before skip=5pt,
  after skip=7pt,
  fontupper=\small\slshape
}

\newcommand{\takeaway}[1]{%
  \begin{takeawaybox}
  \textcolor{caseblue}{\textbf{\textit{Takeaway.}}}\enspace #1
  \end{takeawaybox}
}

\lstdefinestyle{promptstyle}{
  basicstyle=\footnotesize\ttfamily,
  breaklines=true,
  breakatwhitespace=false,
  columns=fullflexible,
  keepspaces=true,
  showstringspaces=false,
  tabsize=2,
  literate={–}{{--}}1 {—}{{---}}1 {…}{{...}}3
           {→}{{$\to$}}1 {↔}{{$\leftrightarrow$}}1
           {─}{{-}}1 {│}{{|}}1 {└}{{+}}1 {├}{{+}}1
           {✓}{{[PASS]}}6 {✗}{{[FAIL]}}6
}

\newtcblisting{promptbox}[2][]{
  enhanced,
  breakable,
  listing only,
  listing engine=listings,
  listing options={style=promptstyle},
  colback=gray!3,
  colframe=gray!70!black,
  colbacktitle=gray!70!black,
  coltitle=white,
  title={#2},
  fonttitle=\bfseries,
  boxrule=0.7pt,
  arc=1.5mm,
  left=4mm,
  right=4mm,
  top=2mm,
  bottom=2mm,
  before skip=8pt,
  after skip=10pt,
  #1
}

\newtcolorbox{caseinput}[1]{
  enhanced, breakable, colback=caseblueback, colframe=caseblue,
  colbacktitle=caseblue, coltitle=white, title={#1}, fonttitle=\bfseries,
  boxrule=0.7pt, arc=1mm, left=4mm, right=4mm, top=2.5mm,
  bottom=2.5mm, before skip=5pt, after skip=4pt
}

\newtcolorbox{casesearch}[1]{
  enhanced, breakable, colback=white, colframe=caseblue,
  colbacktitle=caseblue, coltitle=white, title={#1}, fonttitle=\bfseries,
  boxrule=0.7pt, arc=1mm, left=3mm, right=3mm, top=2.5mm,
  bottom=2.5mm, before skip=5pt, after skip=4pt
}

\newtcolorbox{caseoutput}[1]{
  enhanced, breakable, colback=white, colframe=caseblue,
  colbacktitle=caseblue, coltitle=white, title={#1}, fonttitle=\bfseries,
  boxrule=0.7pt, arc=1mm, left=4mm, right=4mm, top=2.5mm,
  bottom=2.5mm, before skip=5pt, after skip=4pt
}

\newtcolorbox{casecalibration}[1]{
  enhanced, breakable, colback=casegreenback, colframe=casegreen,
  colbacktitle=casegreen, coltitle=white, title={#1}, fonttitle=\bfseries,
  boxrule=0.7pt, arc=1mm, left=4mm, right=4mm, top=2.5mm,
  bottom=2.5mm, before skip=5pt, after skip=4pt
}

\newtcolorbox{revisioncasebox}[2]{
  enhanced, breakable, colback=white, colframe=#2,
  colbacktitle=#2, coltitle=white, title={#1}, fonttitle=\bfseries,
  boxrule=0.8pt, arc=1mm, left=4mm, right=4mm, top=2.5mm,
  bottom=2.5mm, before skip=6pt, after skip=5pt
}

\newcommand{\revisionstage}[2]{%
  \par\medskip
  \noindent\colorbox{#1}{%
    \parbox{\dimexpr\linewidth-2\fboxsep\relax}{\textbf{#2}}%
  }%
  \par\smallskip
}

\newcommand{\casearrow}{%
  \par\nointerlineskip
  \begin{center}
    \vspace{-1mm}
    \textcolor{casegray}{\large$\downarrow$}
    \vspace{-1mm}
  \end{center}
}

\newtcblisting{markdownbox}[1][]{
  listing engine=listings,
  listing options={
    basicstyle=\scriptsize\ttfamily, 
    breaklines=true,              
    breakatwhitespace=true,
    numbers=none,                 
    columns=fullflexible,         
    keepspaces=true,              
    aboveskip=0pt,
    belowskip=0pt
  },
  colback=gray!5!white,           
  colframe=gray!70!black,         
  title={\textbf{Prompt (Markdown)}}, 
  fonttitle=\bfseries,
  listing only,                   
  left=5mm, right=5mm, top=3mm, bottom=3mm, 
  breakable,                      
  enhanced,                       
  overlay={\begin{tcbclipinterior}\fill[gray!5!white] (frame.south west) rectangle ([xshift=0mm]frame.north west);\end{tcbclipinterior}}, 
  #1                              
}

\newtcbinputlisting{\markdownfile}[2][]{
  listing engine=listings,
  listing file={#2}, 
  listing options={
    basicstyle=\scriptsize\ttfamily,
    breaklines=true,
    breakatwhitespace=true,
    numbers=none,
    columns=fullflexible,
    keepspaces=true
  },
  colback=gray!5!white,
  colframe=gray!70!black,
  title={\textbf{Prompt (Markdown)}},
  fonttitle=\bfseries,
  listing only,
  breakable,
  enhanced,
  #1
}

\uselogo{}

\title{\papername{}: Adversarial Solver Calibration for Scaling Learnable Terminal Tasks}

\newcommand{\publicday}{Oct.~27, 2026}

\author[*]{Fanzhe Meng}
\author[*]{Guoxin Chen}
\author[ \hspace{-0.3em}]{Jiale Zhao}
\author[ \hspace{-0.3em}]{Shuang Sun}
\author[ \hspace{-0.3em}]{Yuzhi Lin}
\author[$\dag$]{Wayne Xin Zhao}
\author[$\dag$]{Ruihua Song}
\author[ \hspace{-0.3em}]{Ji-Rong Wen}
\author[$^\dag$]{Kai Jia}

\footerlinks{\resourcebuttons}

\correspondingauthor{\{mengfanzhe16, gx.chen.chn, batmanfly, jiakai0419\}@gmail.com, songruihua\_bloon@outlook.com}

\affil[1]{Gaoling School of Artificial Intelligence, Renmin University of China}
\affil[2]{Independent Researcher}
\affil[3]{AweAI Team\footnote{$^*$Equal Contributions. $^\dag$Corresponding authors. \hfill \textbf{Date:} \publicday.}}

\begin{abstract}
Training terminal agents requires executable and verifiable tasks that are not merely solvable, but appropriately challenging for learning.
Executable validation establishes feasibility, yet does not reveal how a task behaves relative to a given solver setting.
In this paper, we present \papername{}, an autonomous terminal-task synthesis system that uses verified solver behavior to revise candidate tasks through adversarial solver calibration.
Multi-solver calibration targets disagreement within a heterogeneous solver pool, whereas contrastive solver calibration targets a designated strong-pass/weak-fail relation; both operationalize a solver-relative learnable zone anchored in demonstrated solvability.
Using \papername{}, we construct 5,431 calibrated terminal tasks.
Our ablations show that both strategies yield more effective supervision than authoring and validation alone or ordinary single-solver feedback.
Models trained on the full collection achieve 31.84\% and 47.57\% on Terminal-Bench 2.1. The largest improvements over the corresponding base model reach 23.97 percentage points on Terminal-Bench 2.1, 26.54 points on SWE-bench Pro, and 30.04 points on Doc2Repo.
Together, these results support solver-relative learnability as a practical target for constructing effective and transferable agent training data.
\end{abstract}

\makeatletter
\begin{document}

\begingroup
\makeatletter
\renewcommand{\thefootnote}{}
\renewcommand{\@makefnmark}{}
\maketitle
\makeatother
\endgroup

\begin{figure}[H]
    \centering
    \includegraphics[width=\textwidth]{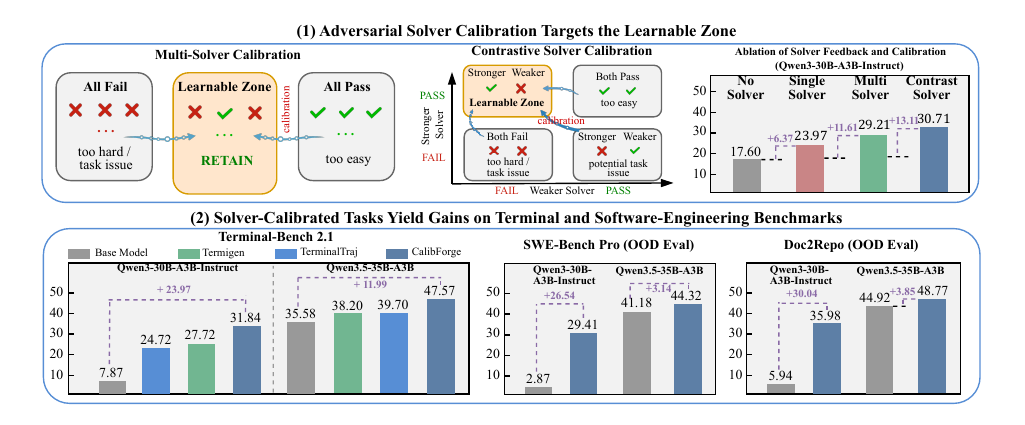}
    \caption{Overview of \papername{}.
    \textbf{Top:} Multi-solver and contrastive calibration target a solver-relative learnable zone through cross-solver disagreement and a strong-pass/weak-fail relation, respectively; both outperform authoring and validation alone and single-solver feedback on Qwen3-30B-A3B-Instruct.
    \textbf{Bottom:} Training on trajectories distilled from 5,431 calibrated tasks improves two Qwen backbones on Terminal-Bench 2.1, SWE-bench Pro, and Doc2Repo.}
    \label{fig:results_overview}
\end{figure}

\section{Introduction}
Large language model agents increasingly tackle complex computing tasks through terminal interfaces~\citep{openai2025codex,anthropic2026claudecode,cursor2026composer2}.
Recent synthesis systems have made substantial progress in constructing and validating terminal tasks at scale~\citep{zhu2026termigen,pi2026data,gandhi2026endless,tang2026agent}.
Yet executable validity alone does not reveal whether a task is appropriately challenging for learning rather than trivial or effectively unsolvable.
Independent attempts by solver agents provide behavioral evidence relevant to this task: their verified outcomes reveal which attempts succeed or fail, while their trajectories help explain why.
This raises a natural question: \textit{how can such solver behavior guide terminal-task construction toward more effective training data?}

Our key idea is to turn task construction into a constrained adversarial author--solver loop, in which solver behavior serves as construction-time feedback.
Solver agents seek successful completion of the current candidate, while the authoring agent uses the resulting behavioral evidence to revise the task toward a desired outcome pattern.
Crucially, a validated candidate is not treated as fixed: the authoring agent may return to external research, reconsider the technical direction, or revise any task component.
Each revised version is revalidated and re-probed before it can be retained.
The loop is constrained by demonstrated solvability, since the target outcome pattern always requires at least one successful solver attempt.
We call this multi-round process \textit{environment-level behavioral calibration}.

We present \textbf{\papername{}}, an autonomous task synthesis system that realizes this process through \textbf{adversarial solver calibration}.
Starting from a clue, \papername{} researches concrete engineering problems, jointly authors the task instruction, execution environment, and verification tests, and validates the resulting candidate through structural checks and self-solving.
It then applies one of two calibration strategies, using solver outcomes and trajectories to guide multi-round revision and re-probing.
\textbf{Multi-solver calibration} targets disagreement within a heterogeneous solver pool, retaining a candidate when at least one solver succeeds and at least one fails.
\textbf{Contrastive solver calibration} targets a strong-pass/weak-fail relation, retaining a candidate when a designated stronger solver succeeds and a designated weaker solver fails.
Together, these criteria operationalize a solver-relative \emph{learnable zone}: retained candidates are demonstrably solvable, yet not uniformly solved under the specified solver setting.

Our experiments show that whether a candidate lies within a solver-relative learnable zone is neither guaranteed by executable validity alone nor fixed at initial construction.
Adversarial solver calibration makes this status observable and actionable: although all candidates entering contrastive calibration have passed structural validation and self-solving, only 19\% exhibit the target relation on the first probe; 96\% are ultimately retained after the revision and re-probing process.
Our ablation shows the downstream value of the learnable zone: multi-solver and contrastive calibration achieve 29.21\% and 30.71\% accuracy on Terminal-Bench 2.1~\citep{merrill2026terminal}, compared with 17.60\% after authoring and validation alone and 23.97\% with single-solver feedback.
Trained on the full collection of 5,431 tasks, \papername{}-30B-A3B and \papername{}-35B-A3B achieve 31.84\% and 47.57\%, surpassing the strongest baselines by 4.12 and 7.87 percentage points.
The benefit extends to out-of-distribution software-engineering benchmarks: the 35B model achieves 44.32\% on SWE-bench Pro~\citep{deng2025swe} for long-horizon issue resolution and 48.77\% on Doc2Repo~\citep{chen2026beyondswe} for full-repository generation, improving over its base model by 3.14 and 3.85 percentage points.

\noindent\begin{minipage}{\linewidth}
Our contributions are summarized as follows:

\begin{itemize}
\item We formulate \textbf{environment-level behavioral calibration}, a construction principle that treats task learnability as solver-relative and uses verified solver behavior to determine retention and guide multi-round revision.

\item We introduce \textbf{\papername{}}, a terminal-task synthesis system that realizes this principle through a constrained adversarial author--solver loop. Its multi-solver and contrastive calibration strategies target cross-solver disagreement and a designated strong-pass/weak-fail relation.

\item We construct \textbf{5,431} calibrated terminal tasks, demonstrate downstream gains over No Solver and Single Solver variants, and show transfer to out-of-distribution SWE benchmarks.
\end{itemize}
\end{minipage}
\par

\begin{figure*}[t]
    \centering
    \includegraphics[width=\linewidth]{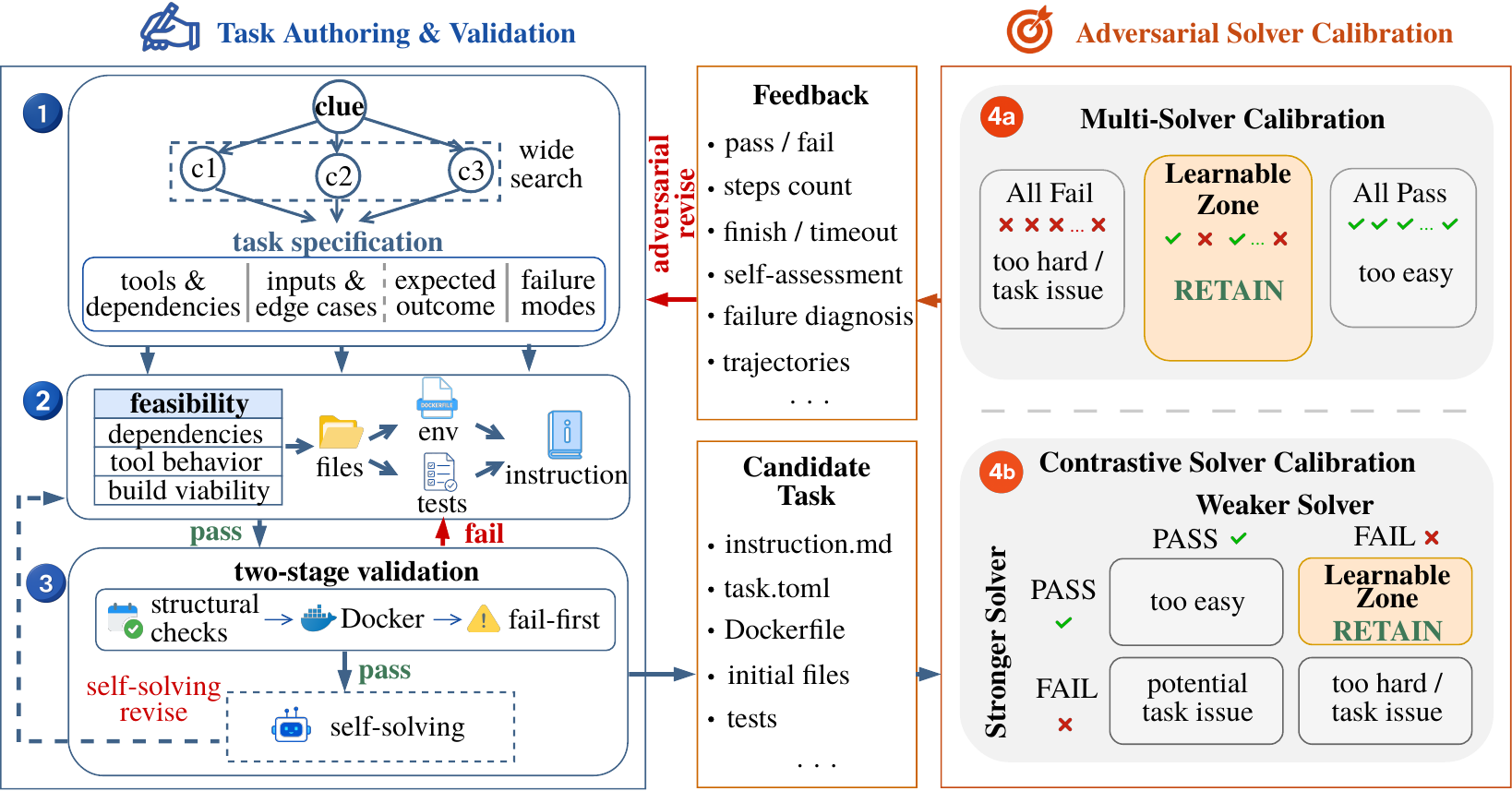}
    \caption{Overview of \papername{}.
    Starting from a \emph{clue}, \papername{} authors a candidate task, subjects it to structural validation and self-solving, and then probes it using one of two adversarial solver-calibration strategies: multi-solver or contrastive calibration.
    The solvers' pass/fail outcomes provide the retention signal, while feedback summaries and full interaction trajectories guide revisions at any stage: the authoring agent may return to web research or revise the task's instruction, environment, or verification tests before revalidation and probing the candidate task again.}
    \label{fig:framework}
\end{figure*}

\begin{minipage}{\linewidth}
\section{Methodology}

\subsection{Overview}
\label{sec:calibforge-overview}
Starting from a clue, \papername{}'s authoring agent constructs a candidate terminal task \(\tau\), validates its structure, and attempts to solve it; \(V(\tau)\) indicates whether both checks pass.
A valid candidate then enters an author--solver loop for \textbf{adversarial solver calibration}, governed by a calibration specification \(\gamma\), which defines the solver setting and behavioral retention criterion \(C_\gamma\).
Each calibration round returns structured feedback summaries containing verified pass/fail outcomes, together with full interaction trajectories.
When the outcomes do not satisfy \(C_\gamma\), the authoring agent uses this feedback to revise and revalidate the task before the next round.
A candidate is eligible for retention when \(C_\gamma\) is satisfied, but the authoring agent may continue revising it.
Any revised candidate must be revalidated and re-probed before retention.
Figure~\ref{fig:framework} illustrates the overall task-construction process, while Algorithm~\ref{alg:calibforge} specifies its iterative control flow.
\end{minipage}
\par

\subsection{Candidate Task Authoring and Validation}
\label{sec:candidate-task-authoring-validation}
\papername{} places the authoring agent in a draft sandbox and equips it with tools for web research, shell command execution, and file editing.
Starting from the clue, the agent searches external technical sources, including official documentation, GitHub repositories and issue discussions, and Stack Overflow questions, for concrete engineering problems such as version-specific bugs, dependency conflicts, configuration pitfalls, and reproducible edge cases.
Based on this evidence, it develops several task directions and selects one based on its relevance to the clue, distinctiveness, and implementation feasibility.
The agent then turns the selected direction into a task specification that records the required tools and dependencies, inputs and edge cases, expected final behaviors, and target failure modes.
Before authoring the task files, it uses the draft sandbox to test whether required packages can be installed, external resources can be accessed, and relevant tools behave as expected, revising the specification when an assumption does not hold.

\begin{algorithm}[tb]
\caption{The Pipeline of \papername{}}
\label{alg:calibforge}
\begin{algorithmic}[1]
\Require clue \(c\), calibration specification \(\gamma\), maximum calibration rounds \(R_{\max}\)

\algphase{tableblue}{I. Task Authoring \& Validation}
\State \(\mathit{spec} \gets \textsc{WideSearchAndSpecify}(c)\)
\State \(\tau \gets \textsc{ConstructTask}(\mathit{spec})\)
\While{\(\neg V(\tau)\)}
    \State \(\tau \gets \textsc{Repair}(\tau)\)
        \Comment{validation failed}
\EndWhile

\algphase{calibrationorange}{II. Adversarial Solver Calibration}
\For{\(r=1,\ldots,R_{\max}\)}
    \State \(\mathit{feedback} \gets \textsc{ProbeAndVerify}(\tau;\gamma)\)
    \State \(\mathit{outcomes} \gets \textsc{Outcomes}(\mathit{feedback})\)
    \If{\(C_{\gamma}(\mathit{outcomes})=1\)}
        \If{the authoring agent ends construction}
            \State \Return \(\tau\)
        \EndIf
    \EndIf
    \State \(\tau \gets \textsc{Revise}(\tau;\mathit{feedback})\)
    \While{\(\neg V(\tau)\)}
        \State \(\tau \gets \textsc{Repair}(\tau)\)
            \Comment{validation failed}
    \EndWhile
\EndFor
\State \Return \textsc{Discard}
\end{algorithmic}
\end{algorithm}

Guided by this specification, the agent jointly constructs the task instruction, an execution environment defined by the Dockerfile and initial files, and verification tests that constitute the candidate task.
Before validation, the agent checks that the initial environment does not expose solution artifacts and that the verification tests match the task instruction without imposing unstated requirements~\citep{bercovich2026makes}.

Before entering solver calibration, each candidate undergoes a two-stage validation process comprising structural validation and self-solving.
During structural validation, \papername{} checks that the required build-context files are present, builds and initializes the execution environment, runs the verifier, and confirms that all tests fail in the initial state.
Once the structural checks pass, \papername{} instantiates an isolated test sandbox from the candidate Dockerfile.
Within this sandbox, the authoring agent attempts to solve the task and runs the verifier against the resulting environment state, checking that the intended solution is executable and that the instruction, environment, and verification tests are mutually consistent.
If either stage fails, the authoring agent repairs the candidate and reevaluates it through both stages.
Accordingly, \(V(\tau)=1\) if and only if the candidate passes both structural validation and self-solving; only such candidates proceed to adversarial solver calibration.

\subsection{Adversarial Solver Calibration}
\label{sec:adversarial-solver-calibration}
Rather than using solver behavior only to evaluate a finished task, \papername{} uses it to revise the candidate toward satisfying \(C_\gamma\).

\paragraph{Calibration Loop.}
In each calibration round, \papername{} instantiates an isolated sandbox from the candidate's Dockerfile for every solver subagent, assigns the solvers according to \(\gamma\), and provides each with the same task instruction.
After an attempt, \papername{} runs the task verifier against the final sandbox state, producing a verified outcome \(y_i\in\{0,1\}\), where \(y_i=1\) indicates that all verification tests pass.
Each attempt returns a structured feedback summary containing the verified pass/fail outcome, step count, completion status, self-assessment, and failure diagnosis, together with the full interaction trajectory.
The pass/fail outcomes determine whether the candidate satisfies \(C_{\gamma}\), while the feedback summary and full interaction trajectories guide task revision.

\paragraph{Multi-Solver Calibration.}
Under \(\gamma_{\mathrm{multi}}\), \papername{} dispatches \(K\) solver subagents using different models to attempt the same candidate independently.
Let \(\mathbf{y}=(y_1,\ldots,y_K)\) denote their verified outcomes.
The retention criterion requires disagreement across solver models: at least one attempt must pass, while at least one must fail.
Formally, the multi-solver retention criterion is
\begin{equation}
C_{\mathrm{multi}}(\mathbf{y})
=
\mathbf{1}\!\left[\,0<\sum_{i=1}^{K}y_i<K\,\right].
\label{eq:calib-multi}
\end{equation}
An all-pass round may indicate that the task is too easy, for example because it admits a shallow solution path, whereas an all-fail round may indicate excessive difficulty, underspecification, or a broken task.
The feedback summaries and trajectories help the authoring agent diagnose these outcome patterns and revise the task before another calibration round.
By using disagreement across solver models as its retention criterion, multi-solver calibration reduces reliance on any single solver and captures a broader range of solution paths and failure modes.

\paragraph{Contrastive Solver Calibration.}
Under \(\gamma_{\mathrm{con}}\), \papername{} runs a designated stronger solver and a designated weaker solver on the same candidate.
Writing \(y_{\mathrm{s}}\) and \(y_{\mathrm{w}}\) for their verified outcomes, respectively, \(C_{\gamma}\) is instantiated as
\begin{equation}
C_{\mathrm{con}}(y_{\mathrm{s}},y_{\mathrm{w}})
=
\mathbf{1}\!\left[\,y_{\mathrm{s}}=1\land y_{\mathrm{w}}=0\,\right].
\label{eq:calib-contrast}
\end{equation}
The weaker solver's failure shows that the task lies beyond the weaker setting, while the stronger solver's success confirms that it remains solvable under the stronger setting; together, these outcomes place the candidate within the capability interval defined by the two solver settings.
If both solvers pass, the authoring agent searches for shortcuts or insufficient difficulty; if both fail, it checks solvability, specification quality, and verification; if the weaker solver passes while the stronger solver fails, the authoring agent inspects the task for leakage, nondeterminism, or misleading formulation.

The specifications \(\gamma_{\mathrm{multi}}\) and \(\gamma_{\mathrm{con}}\) therefore instantiate \(\gamma\) with different solver settings and retention criteria while sharing the same author--solver loop.

\section{Experiments}
\subsection{Experimental Setup}
\label{subsec:setup}

\textbf{Task Construction and Calibration.}
\papername{} uses DeepSeek-V4-Pro~\citep{xu2026deepseek} as the authoring agent.
For multi-solver calibration, three solver subagents using DeepSeek-V4-Flash, GLM-5~\citep{zeng2026glm}, and Kimi K2.5~\citep{team2026kimi}, attempt each candidate independently.
For contrastive solver calibration, DeepSeek-V4-Pro and DeepSeek-V4-Flash serve as the designated stronger and weaker solvers, respectively.
Each solver attempt is limited to 100 interaction steps and 30 minutes.
For each candidate, the author--solver loop runs for at most \(R_{\max}=50\) calibration rounds.

\textbf{Trajectory Collection and Training.}
Retained tasks are stored as Harbor-style task instances~\citep{Harbor_Framework}.
We distill SFT trajectories with DeepSeek-V4-Pro (reasoning effort \texttt{high}) under \textsc{\papername{}-Eval}, a minimal code-agent scaffold that exposes only bash, file-editing, and finish tools, with a design similar to the code-task evaluation framework described in DeepSeek-V4~\citep{xu2026deepseek}.
For both \papername{} and baseline task sets, we collect two attempts per task using the same teacher model, each limited to 200 steps and one hour.
We retain test-passing trajectories, then filter them for length, invalid tool calls, and tokenizer-unsafe special tokens.
We fine-tune Qwen3-30B-A3B-Instruct and Qwen3.5-35B-A3B~\citep{qwen3technicalreport,qwen3.5} using full-parameter SFT and train for 10 epochs.

\textbf{Baselines.}
We compare with open-source terminal-task synthesis methods that release their task sets: Endless Terminals, SETA-Env, CLI-Gym, TermiGen, and TerminalTraj~\citep{gandhi2026endless,setaenv2026,lin2026cli,zhu2026termigen,wu2026large}.
For SETA-Env, we use its January 2026 public release containing 1,375 tasks.
For each baseline, we re-distill its full released task set using the same teacher protocol and apply the same training recipe.
On Qwen3.5-35B-A3B, we retrain TermiGen and TerminalTraj, the two best-performing baselines on Qwen3-30B-A3B-Instruct.

\textbf{Benchmarks and Evaluation.}
We use Terminal-Bench 2.1~\citep{merrill2026terminal} as the primary benchmark, evaluating it with a 500-step limit and the same one-hour per-task timeout used for trajectory collection.
Each Terminal-Bench evaluation sandbox is capped at 16 CPUs and 48 GiB RAM, with internet access enabled for dependency installation and access to external resources, consistent with the benchmark design.
To evaluate out-of-distribution transfer, we additionally use the 731-task public set of SWE-bench Pro~\citep{deng2025swe} for complex, long-horizon software engineering in existing repositories and Doc2Repo~\citep{chen2026beyondswe,zhao2026denovoswe} for full-repository generation from natural-language specifications.
Both are evaluated using their official scaffolds in sandboxes capped at 4 CPUs and 8 GiB RAM, with internet access disabled during agent execution.
After evaluation, we audit all successful trajectories for solution lookup and verifier manipulation.
We report task accuracy on Terminal-Bench 2.1 and SWE-bench Pro, and \emph{Pass Rate} on Doc2Repo.
For SFT evaluations, Terminal-Bench 2.1 and Doc2Repo results are reported as mean \(\pm\) standard error of the mean (SEM) over three runs, while SWE-bench Pro is evaluated once due to its higher evaluation cost.

\textbf{Benchmark Decontamination.}
We decontaminate the training tasks against all three evaluation benchmarks.
Following prior work~\citep{pi2026data}, we first remove candidates whose prompts have 14-gram overlap with any evaluation instance.
We then compute 5-shingle Jaccard similarity over normalized instructions and available verifier or test code, and combine it with shared output paths, overlapping test functions, and high-risk task-family matches.
Any candidate flagged as overlapping with any of the three benchmarks is discarded; the supplementary material gives the full matching rule.

\subsection{Main Results}

\begin{table}[t]
    \centering
    \small
    \setlength{\tabcolsep}{7pt}
    \begin{tabular}{@{}lc>{\columncolor{tableblue}}c>{\columncolor{tableblue}}c@{}}
    \toprule
    \textbf{Training Data} &
    \makecell[c]{\textbf{TB2.1}\\\textbf{Acc. (\%)}} &
    \makecell[c]{\textbf{SWE-Pro}\\\textbf{Resolved (\%)}} &
    \makecell[c]{\textbf{Doc2Repo}\\\textbf{Pass Rate (\%)}} \\
    \midrule
    \multicolumn{4}{c}{\textit{\textbf{Qwen3-30B-A3B-Instruct}}} \\
    \midrule
    Base Model & \msem{7.87}{0.00} & 2.87 & \msem{5.94}{0.88} \\
    Endless Terminals & \msem{18.73}{2.09} & 21.34 & \msem{18.26}{2.39} \\
    CLI-Gym & \msem{22.85}{1.63} & \underline{28.32} & \msem{29.91}{1.53} \\
    SETA-Env & \msem{22.85}{1.50} & 26.40 & \msem{24.91}{1.84} \\
    TerminalTraj & \msem{24.72}{1.12} & 25.72 & \msem{24.36}{0.96} \\
    TermiGen & \underline{\msem{27.72}{0.99}} & 26.81 & \underline{\msem{34.11}{2.64}} \\
    \textbf{\papername{}} & \textbf{\msem{31.84}{1.35}} & \textbf{29.41} & \textbf{\msem{35.98}{1.82}} \\
    \midrule
    \multicolumn{4}{c}{\textit{\textbf{Qwen3.5-35B-A3B}}} \\
    \midrule
    Base Model & \msem{35.58}{0.37} & 41.18 & \msem{44.92}{1.14} \\
    TermiGen & \msem{38.20}{1.72} & 43.37 & \msem{44.54}{1.58} \\
    TerminalTraj & \underline{\msem{39.70}{0.99}} & \underline{43.91} & \underline{\msem{47.20}{1.89}} \\
    \textbf{\papername{}} & \textbf{\msem{47.57}{0.37}} & \textbf{44.32} & \textbf{\msem{48.77}{0.90}} \\
    \bottomrule
    \end{tabular}
    \caption{Main results on Terminal-Bench 2.1 and out-of-distribution software-engineering benchmarks. \colorbox{tableblue}{\strut Blue}-shaded SWE-bench Pro and Doc2Repo columns denote out-of-distribution evaluations. Results are reported as mean \(\pm\) SEM over three runs except for SWE-bench Pro, which is evaluated once. \textbf{Bold} and \underline{underlined} denote the best and second-best result within each backbone block.}
    \label{tab:main_results}
\end{table}

\begin{figure*}[!t]
\centering
\begin{subfigure}[t]{0.48\textwidth}
    \centering
    \includegraphics[width=0.90\linewidth]{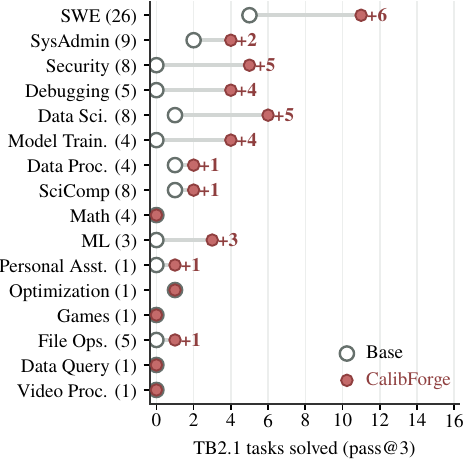}
    \caption{\papername{}-30B-A3B}
    \label{fig:per_category_30b}
\end{subfigure}\hfill
\begin{subfigure}[t]{0.48\textwidth}
    \centering
    \includegraphics[width=0.90\linewidth]{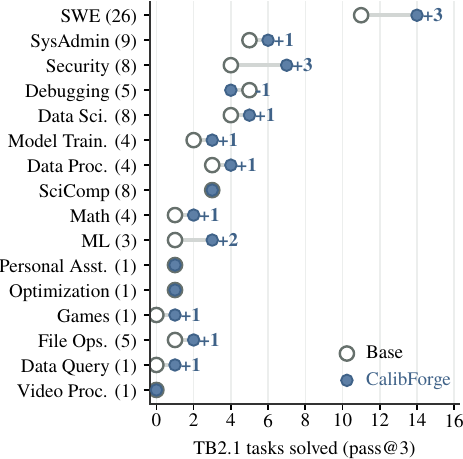}
    \caption{\papername{}-35B-A3B}
    \label{fig:per_category_35b}
\end{subfigure}
\caption{Per-category tasks solved for (a) \papername{}-30B-A3B and (b) \papername{}-35B-A3B, compared with their respective base backbones. Each row represents a task category, with its total number of benchmark tasks shown in parentheses. Markers indicate the number of distinct tasks solved at least once across the included evaluation runs. Signed annotations report the change in solved-task counts relative to the base model. SWE, SysAdmin, SciComp, and ML denote software engineering, system administration, scientific computing, and machine learning, respectively.}
\label{fig:per_category}
\end{figure*}

\textbf{Terminal-Task Performance.}
On Terminal-Bench 2.1, \papername{}-30B-A3B and \papername{}-35B-A3B achieve 31.84\% and 47.57\%, outperforming the strongest baselines under the shared training protocol by 4.12 and 7.87 percentage points.
Because all task sets use the same distillation and training protocol, the training-data source is the primary experimental difference.
Figure~\ref{fig:per_category} further shows gains across a broad range of task categories.
Both models improve or match their respective base models in most categories.

\textbf{Cross-Benchmark Transfer.}
On SWE-bench Pro, \papername{}-30B-A3B and \papername{}-35B-A3B improve over their respective base models by 26.54 and 3.14 percentage points; on Doc2Repo, they improve by 30.04 and 3.85 points, respectively.

\takeaway{Across two backbones, \papername{} tasks provide more effective supervision than the baseline datasets.
Their gains on repository-level generation and complex SWE tasks further suggest that the learned capabilities transfer beyond the terminal-task distribution used for training.}

\subsection{Analysis of Synthesized Tasks}

\begin{figure*}[!t]
\centering
\includegraphics[width=\linewidth]{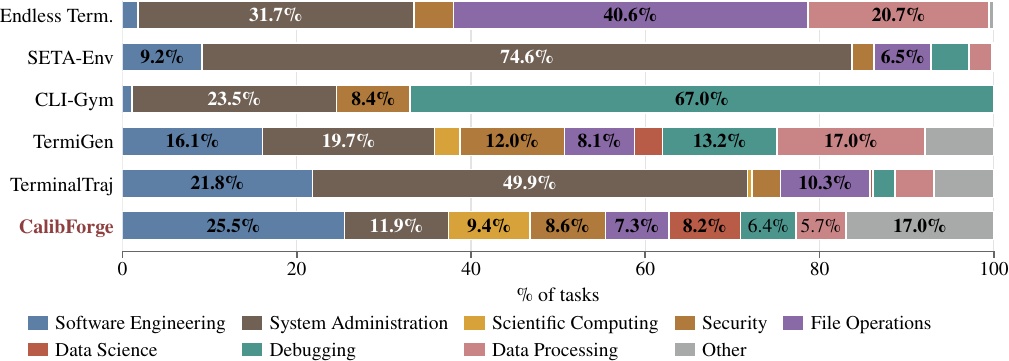}
\caption{Domain composition of tasks synthesized by \papername{} and prior terminal-task datasets under a common 16-category taxonomy.
Categories accounting for at least 5\% of tasks synthesized by \papername{} are shown separately, with the remainder grouped as \emph{Other}.}
\label{fig:domain_comparison}
\end{figure*}

\textbf{Scale and Domain Coverage.}
The resulting 5,431 tasks comprise 1{,}263 tasks from multi-solver calibration and 4{,}168 from contrastive solver calibration.
The tasks span all 16 categories in our domain taxonomy.
To characterize domain coverage, Figure~\ref{fig:domain_comparison} compares tasks synthesized by \papername{} with prior terminal-task sets under a common taxonomy.
Several prior datasets are dominated by one or two categories: system administration accounts for 74.6\% of SETA-Env and 49.9\% of TerminalTraj, debugging accounts for 67.0\% of CLI-Gym, and file operations accounts for 40.6\% of Endless-Terminals.
In contrast, the largest category in the final \papername{} task set is software engineering at 25.5\%, with substantial coverage also allocated to system administration, scientific computing, security, file operations, data science, debugging, and data processing.
This composition indicates that the synthesized tasks broaden terminal supervision beyond the narrow domain concentrations present in several existing task sets.

\begin{figure}[t]
    \centering
    \includegraphics[width=0.5\linewidth]
    {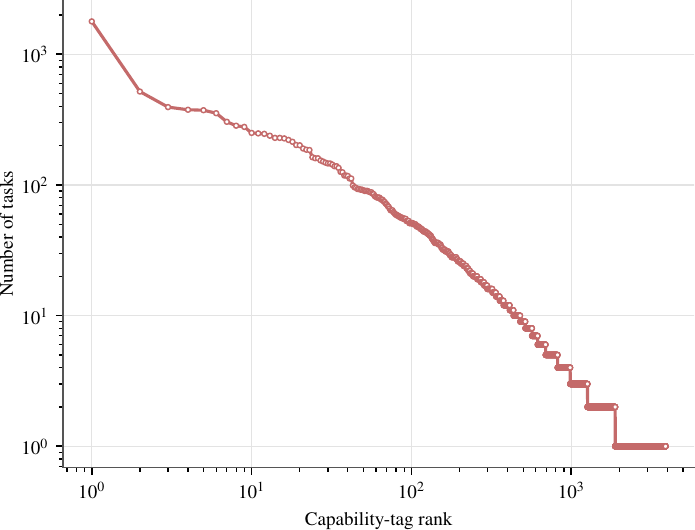}
    \caption{Rank--frequency distribution of capability tags in the \papername{} task collection.}
    \label{fig:capability-long-tail}
\end{figure}

\textbf{Capability Diversity.}
Beyond broad domain categories, we examine diversity in the tools, techniques, and problem-solving capabilities exercised by individual tasks.
Following the Terminal-Bench 2.0 annotation scheme~\citep{merrill2026terminal}, we represent these properties as capability tags.
As shown in Figure~\ref{fig:capability-long-tail}, we observe 3,885 distinct capability tags, with a median of five tags per task.
Their rank--frequency distribution is long-tailed: 51.6\% of the distinct tags occur in only one task, and 82.2\% occur in at most five tasks.
This long tail shows that the domain breadth is accompanied by many specialized capabilities rather than repeated combinations of a small capability vocabulary.

\begin{figure}[t]
    \centering
    \captionsetup[subfigure]{skip=2pt}

    \begin{subfigure}[t]{0.45\linewidth}
        \centering
        \includegraphics[width=\linewidth]
        {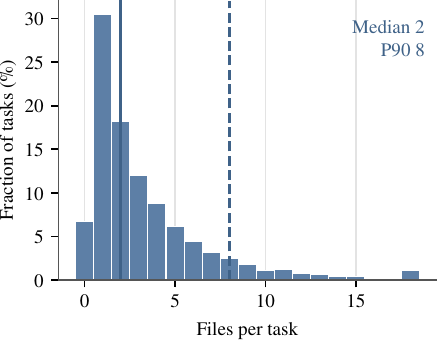}
        \caption{Initial artifacts}
        \label{fig:initial-artifacts}
    \end{subfigure}
    \hfill
    \begin{subfigure}[t]{0.45\linewidth}
        \centering
        \includegraphics[width=\linewidth]
        {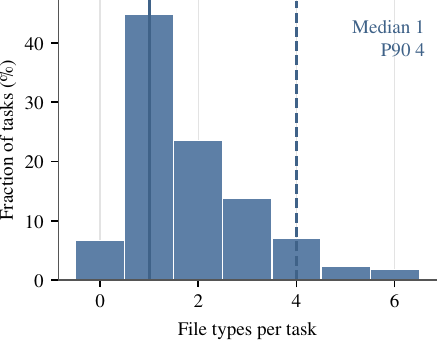}
        \caption{Distinct file types}
        \label{fig:file-types}
    \end{subfigure}

    \medskip

    \begin{subfigure}[t]{0.45\linewidth}
        \centering
        \includegraphics[width=\linewidth]
        {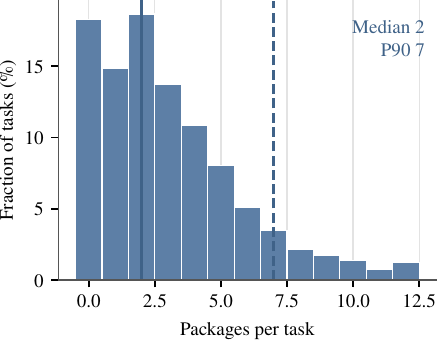}
        \caption{Environment dependencies}
        \label{fig:environment-dependencies}
    \end{subfigure}
    \hfill
    \begin{subfigure}[t]{0.45\linewidth}
        \centering
        \includegraphics[width=\linewidth]
        {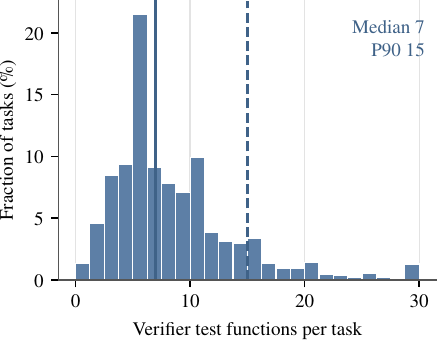}
        \caption{Verifier test functions}
        \label{fig:verifier-tests}
    \end{subfigure}

    \caption{Per-task distributions of environment and verification statistics.
    Solid and dashed vertical lines denote the median and 90th percentile, respectively.}
    \label{fig:task-structure}
\end{figure}

\textbf{Environment and Verification Statistics.}
Figure~\ref{fig:task-structure} presents the per-task distributions of initial artifacts, distinct file types, environment dependencies, and verifier test functions, while Table~\ref{tab:task-statistics} summarizes the corresponding statistics.
A median task contains two initial artifacts, one distinct file type, two environment dependencies, and seven verifier test functions; the corresponding 90th-percentile values are eight, four, seven, and 15.
These distributions show that the collection varies not only in task topic, but also in the amount of environment state and verification logic that a solver must handle.

\begin{table}[t]
    \centering
    \small
    \setlength{\tabcolsep}{5pt}
    \begin{tabular}{@{}lrrrr@{}}
        \toprule
        \textbf{Metric} &
        \textbf{Collection} &
        \textbf{Median} &
        \textbf{IQR} &
        \textbf{P90} \\
        \midrule
        \rowcolor{tableblue}
        \multicolumn{5}{l}{\textbf{Environment}} \\
        Initial artifacts
            & 19,911 & 2 & 1--4 & 8 \\
        Distinct file types
            & 362 & 1 & 1--2 & 4 \\
        Distinct environment dependencies
            & 615 & 2 & 1--4 & 7 \\
        \rowcolor{tableblue}
        \multicolumn{5}{l}{\textbf{Verification}} \\
        Verifier test functions
            & 45,953 & 7 & 5--11 & 15 \\
        \bottomrule
    \end{tabular}
    \caption{Collection-level and per-task statistics of environment and verification components.
    IQR denotes the interval between the first and third quartiles; P90 denotes the 90th percentile.}
    \label{tab:task-statistics}
\end{table}

\begin{figure}[H]
\centering
\begin{subfigure}[t]{0.49\linewidth}
    \centering
    \includegraphics[width=\linewidth]{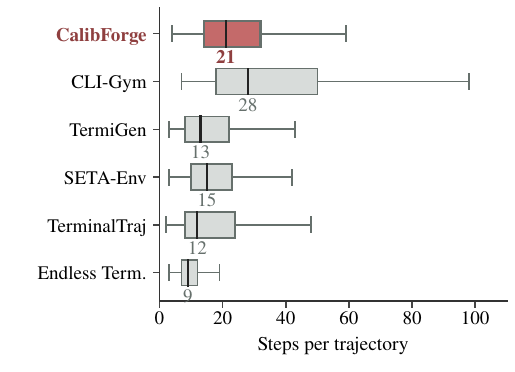}
    \caption{Interaction steps}
    \label{fig:traj_steps}
\end{subfigure}\hfill
\begin{subfigure}[t]{0.49\linewidth}
    \centering
    \includegraphics[width=\linewidth]{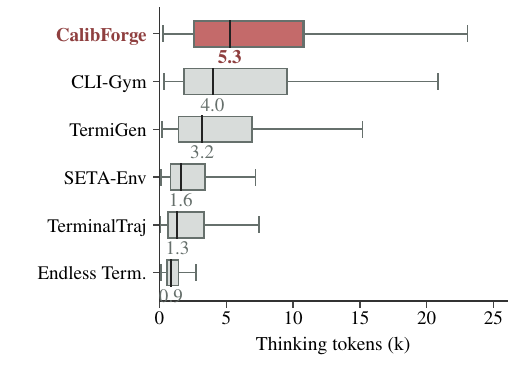}
    \caption{Teacher thinking tokens}
    \label{fig:traj_think}
\end{subfigure}
\caption{Trajectory characteristics across task sets re-distilled under the identical teacher protocol: (a) interaction steps and (b) teacher thinking tokens.
Numbers below the boxes denote medians.
CLI-Gym produces the longest trajectories by interaction count, whereas \papername{} elicits the largest teacher reasoning budget.}
\label{fig:traj_depth}
\end{figure}

\takeaway{Terminal-task diversity is multidimensional: domain coverage, capability variety, and environment and verification structure capture complementary properties of a training collection.
Task count or domain breadth alone is therefore insufficient to characterize the range of behaviors represented by synthesized terminal tasks.}

\textbf{Trajectory Characteristics.}
We next compare per-trajectory interaction steps and teacher thinking tokens across task sets re-distilled under the identical teacher protocol (Figure~\ref{fig:traj_depth}).
CLI-Gym derives repository-grounded environment-repair tasks by inverting runnable Python project environments~\citep{lin2026cli}.
Such tasks typically require dependency or configuration diagnosis followed by repeated execution and verification, so their task structure naturally produces longer interaction traces.
Consistent with this task structure, CLI-Gym has the highest median interaction length at 28 steps, whereas \papername{} has a median of 21 steps; the ordering reverses for thinking tokens, where \papername{} has the highest median at 5.3k compared with 4.0k for CLI-Gym.
Step count alone is therefore insufficient to characterize the reasoning depth or quality of a terminal trajectory.

\FloatBarrier

\subsection{Effect of Solver Calibration}

\textbf{Ablation Variants.}
\emph{No Solver} omits external solver feedback.
\emph{Single Solver} uses a separate solver subagent powered by the same DeepSeek-V4-Pro model as the authoring agent.
The authoring agent receives the solver's pass/fail outcome and execution trajectory and decides whether and how to revise the candidate based on the observed solver behavior, without a prescribed target outcome pattern.
This variant tests whether solver feedback alone provides comparable benefits to feedback guided by an explicit calibration criterion.
\emph{Multi Solver} and \emph{Contrastive Solver} instantiate the two retention criteria introduced in the Methodology section.

\begin{table}[t]
\centering
\normalsize
\setlength{\tabcolsep}{3pt}
\begin{tabular}{@{}lcc@{}}
\toprule
\makecell[c]{\textbf{Calibration}\\\textbf{Mode}} &
\makecell[c]{\textbf{SFT}\\\textbf{Trajectories}} &
\makecell[c]{\textbf{TB2.1}\\\textbf{Acc. (\%)}} \\
\midrule
No Solver & 2,466 & \msem{17.60}{0.99} \\
Single Solver & 2,493 & \msem{23.97}{1.98} \\
\midrule
\rowcolor{tableblue}
Multi Solver & 2,425 & \msem{29.21}{1.12} \\
\rowcolor{tableblue}
Contrastive Solver & 2,561 & \textbf{\msem{30.71}{2.28}} \\
\bottomrule
\end{tabular}
\caption{Effect of solver feedback and behavioral calibration on Qwen3-30B-A3B-Instruct.
Blue-shaded rows denote our multi-solver and contrastive calibration strategies.
Each variant contains 1,300 tasks and uses the same teacher-distillation and SFT recipe; SFT trajectory counts reflect trajectory filtering.}
\label{tab:solver_ablation}
\end{table}

\textbf{Adversarial solver calibration provides larger gains than single-solver feedback.}
Table~\ref{tab:solver_ablation} compares four variants constructed from the same set of source clues, each containing 1,300 tasks.
Of the 1,300 multi-solver tasks used in the ablation, 1,263 were retained in the final collection after additional filtering (Appendix~\ref{supp:sft-details}).
Relative to No Solver, single-solver feedback improves Terminal-Bench 2.1 accuracy by 6.37 points, while multi-solver and contrastive calibration yield gains of 11.61 and 13.11 points, respectively.
Multi-solver calibration achieves higher accuracy despite retaining fewer training trajectories than No Solver.
\takeaway{Under the same clue settings, both calibration strategies outperform single-solver feedback.
These results support guiding task revision with explicit solver-relative criteria rather than leaving the authoring agent to interpret solver feedback without a prescribed target outcome pattern.}

\begin{figure}[!htbp]
\centering
\includegraphics[width=0.76\linewidth]{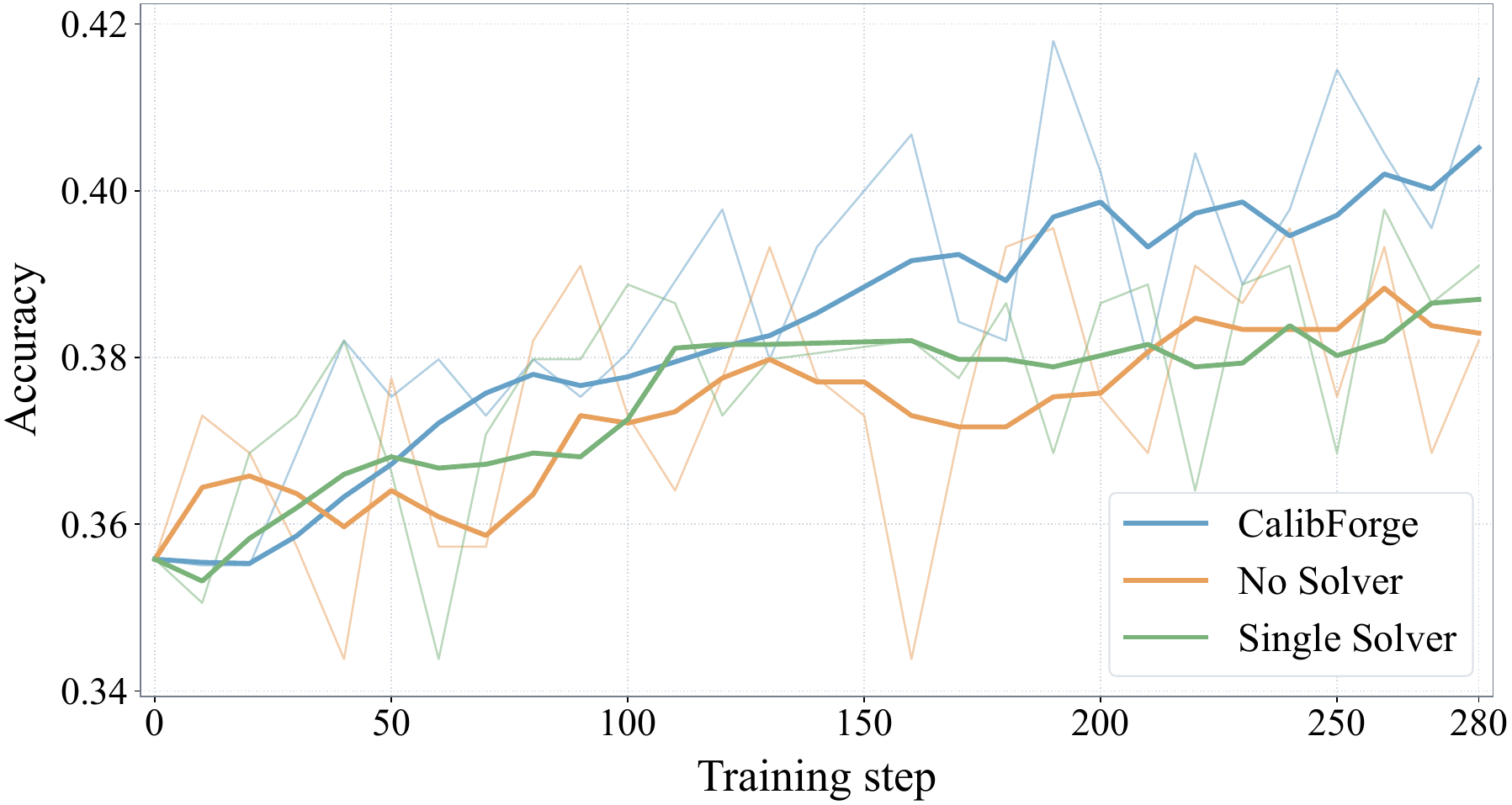}
\caption{GRPO on 1,024-task subsets with Qwen3.5-35B-A3B.
Thin lines show TB2.1 accuracy averaged over five runs;
thick lines show trailing five-checkpoint means
(shorter windows initially).}
\label{fig:grpo_ablation}
\end{figure}

\textbf{Reinforcement Learning on Synthesized Tasks.}
We train Qwen3.5-35B-A3B using GRPO on three sets of 1,024 tasks each.
The \papername{} set is sampled from the combined multi-solver and contrastive-solver collections, while the other two sets are sampled from the No Solver and Single Solver ablations.
We sample 16 rollouts per task, use verifier-based rewards of $+1$ for success and $-1$ for failure, and normalize advantages within each group.
For each evaluated checkpoint, we report the mean Terminal-Bench 2.1 accuracy over five evaluation runs.
Figure~\ref{fig:grpo_ablation} shows that \papername{} generally achieves higher accuracy than the No Solver and Single Solver variants over the first 280 training steps, with a clearer advantage in the later stages.
This comparison suggests that solver-calibrated tasks provide more effective training signals for GRPO than tasks constructed without calibration or with single-solver feedback alone.

We next analyze contrastive-calibration dynamics, tracking candidates from their initial solver outcomes to final retention.

\begin{figure}[!t]
\centering
\includegraphics[width=0.5\textwidth]{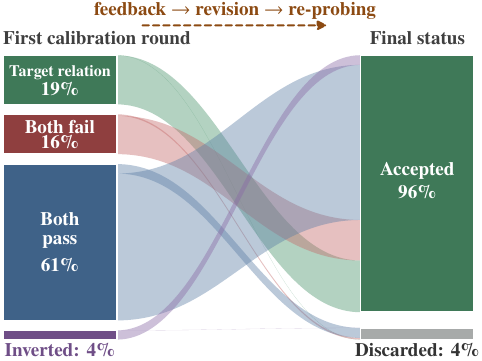}
\caption{First-probe solver outcomes and final retention under contrastive calibration.
The left nodes classify the first verified solver probe: 19\% exhibit the target strong-pass/weak-fail relation.
The right nodes report final run-level retention, which reaches 96\% after revision and re-probing.}
\label{fig:contrast_outcome_flow}
\end{figure}

\textbf{Beyond task validity, solver feedback revises rather than merely filters candidates.}
Figure~\ref{fig:contrast_outcome_flow} classifies each candidate by its first verified solver probe: only 19\% exhibit the target strong-pass/weak-fail relation, while 81\% do not, despite all candidates having passed structural validation and self-solving.
The dominance of both-pass outcomes shows that, in this setting, the most common gap left by validity checks is insufficient separation between the solver settings rather than task unsolvability.
Structural validation and self-solving establish executability and solvability, but do not control where a task lies within the target capability interval defined by the two solver settings.
After feedback-driven revision and re-probing, 96\% of candidates are ultimately retained.

\begin{figure}[!t]
\centering
\includegraphics[width=0.46\columnwidth]{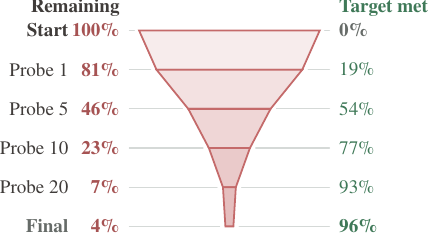}
\caption{Progressive contrastive-calibration funnel.
Width shows the fraction of candidates yet to meet the target relation.
Right-hand labels show the cumulative percentage meeting it by each probe count.}
\label{fig:contrast_acceptance_curve}
\end{figure}

\textbf{Feedback-driven revision corrects many mismatches early, while harder candidates require sustained calibration.}
Figure~\ref{fig:contrast_acceptance_curve} shows the cumulative fraction of candidates that have exhibited the target strong-pass/weak-fail relation by each probe count.
The target relation is observed on the first probe for 19\% of candidates, within five probes for 54\%, and within twenty probes for 93\%.
The early gains show that many mismatches can be corrected with few revisions, while the long tail shows that harder cases require sustained probing.
The calibration budget therefore affects not only construction cost but also which recoverable candidates enter the training set: a short horizon favors candidates whose mismatches can be corrected with only a few revisions.
\takeaway{A non-target solver outcome is often a correctable mismatch rather than an intrinsic defect in the candidate.
Treating such outcomes as diagnostic evidence allows calibration to revise recoverable tasks toward the target relation, while the long tail shows that some mismatches require sustained calibration rather than only a few revisions.}

\FloatBarrier
\Needspace{8\baselineskip}
\section{Related Work}
\label{sec:related_work}

\textbf{Terminal Agents and Benchmarks.}
Execution-grounded agents operate in interactive software environments, spanning code and repository tasks as well as stateful command-line workspaces~\citep{yang2023intercode,jimenez2024swe,yang2024swe,yao2022react,liu2024agentbench,xie2024osworld,siegel2024core,song2026swe,sun2026swe,bui2026building,ren2026self,ivison2026tmax,wang2025openhands,chen2026toward,cheng2026llm}.
Terminal-Bench, TerminalWorld, and OpenThoughts-TBLite evaluate terminal agents using executable tasks and verifiers~\citep{merrill2026terminal,chu2026terminalworld,OpenThoughts-TBLite,feng2026longcli,chen2026tua,li2026long}.
Such tasks couple instructions, initial files, dependencies, and verification tests, making their joint construction and validation essential.

\textbf{Verifiable Terminal-Task Synthesis.}
Existing methods synthesize terminal tasks from specifications and capability taxonomies~\citep{zhu2026termigen,pi2026data,gandhi2026endless,shen2026seta,hua2026cli,peng2026litecoder,lai2026clawforge,zhao2026nexforge,chen2026expanding}, from software artifacts, repositories, and agent trajectories~\citep{lin2026cli,wu2026large,yang2026makes,zhao2026immersion}, or from structured agent skills and skill graphs~\citep{cheng2026terminal,fan2026toward}.
These methods combine environment construction with execution-grounded validation, whereas \papername{} studies how solver behavior can guide candidate revision.

\textbf{Behavioral Feedback for Data Construction.}
Prior work uses model behavior to adapt benchmarks or curricula~\citep{kiela2021dynabench,dennis2020emergent}, while Reflexion uses trajectory feedback to improve a solver's subsequent attempts~\citep{shinn2023reflexion}.
\papername{} instead applies solver behavior to the task under construction: verified pass/fail outcomes determine retention, while feedback summaries and full interaction trajectories guide task revision.

\section{Conclusion}

We presented \papername{}, an autonomous terminal-task synthesis system that turns solver behavior into construction-time feedback through two adversarial solver-calibration strategies: multi-solver and contrastive calibration.
Fine-tuning two Qwen backbones on trajectories distilled from 5,431 calibrated tasks yields gains on Terminal-Bench 2.1 and transfer to repository-level software-engineering benchmarks.
Matched ablations attribute these gains to the two calibration strategies rather than trajectory volume or single-solver feedback.
These results show that solver behavior can guide the construction of more effective terminal-task training data.

\FloatBarrier
\section*{Limitations}

In addition to the checks provided by \papername{}'s validation and solver calibration, we manually inspected selected synthesized tasks.
These checks do not amount to exhaustive human review of the full training collection, so errors in task specifications or verifiers may remain.

Calibration also involves a cost--reliability trade-off.
To limit calibration costs, we use one attempt per solver in each round, so the observed pass/fail relation may vary across repeated attempts on the same task.
Estimating solver success rates from multiple attempts could provide a more robust signal for task revision and retention, at additional computational cost.

\section*{Ethical Considerations}

Agents trained on the synthesized tasks may acquire system-administration and security capabilities that could be misused for unauthorized operations.
Their deployment should therefore include human oversight, resource limits, and access controls for sensitive resources.

\clearpage
\bibliography{main}
\clearpage
\appendix

\makeatletter
\@addtoreset{figure}{section}
\@addtoreset{table}{section}
\makeatother
\renewcommand{\thefigure}{\Alph{section}\arabic{figure}}
\renewcommand{\thetable}{\Alph{section}\arabic{table}}

\providecommand{\casecontinuedtitle}{Case continued}
\renewcommand{\revisionstage}[2]{%
  \par\medskip
  \noindent\colorbox{#1}{%
    \parbox{\dimexpr\linewidth-2\fboxsep\relax}{\textbf{#2}}%
  }%
  \par\nobreak\smallskip\nobreak
}
\renewtcolorbox{revisioncasebox}[2]{
  enhanced, breakable, colback=white, colframe=#2,
  colbacktitle=#2, coltitle=white, title={#1},
  fonttitle=\bfseries\raggedright, title after break={\casecontinuedtitle},
  boxrule=0.8pt, arc=1mm, left=4mm, right=4mm, top=2.5mm,
  bottom=2.5mm, before skip=6pt, after skip=5pt
}
\raggedbottom
\tcbset{lines before break=2, ignore nobreak}

\section{From a Clue to a Calibrated Task}
\label{supp:clue-to-task}

\paragraph{Use of External Resources.}
During task authoring, \papername{} may consult publicly available code repositories and technical documentation for task ideas and implementation references.
These resources inform the construction of task instructions, executable environments, and programmatic verifiers.
The released data have been anonymized to remove personally identifying information.

The following example traces how \papername{} uses web research to develop a clue into a candidate terminal task and retains it through contrastive solver calibration.

\begin{caseinput}{\textcircled{\scriptsize 1}\quad Input $c$ --- Clue}
{\small
\begin{tabularx}{\linewidth}{@{}>{\raggedright\arraybackslash\bfseries}p{0.30\linewidth}X@{}}
Domain & Data Processing \& ETL \\
Scope & Pipelines, transformation, schema evolution, and log analysis \\
Capability hints & Data I/O; Binary Parsing; Data Recovery \\
Tool hints & \texttt{xxd}; \texttt{hexdump} \\
\end{tabularx}
}
\end{caseinput}

\casearrow

\begin{casesearch}{\textcircled{\scriptsize 2}\quad Wide Search and Direction Selection}
\small
The authoring agent issued \textbf{24 search calls} spanning the major directions below.

\medskip
\renewcommand{\arraystretch}{1.15}
\begin{tabularx}{\linewidth}{@{}>{\raggedright\arraybackslash\bfseries}p{0.20\linewidth}
                                >{\raggedright\arraybackslash}p{0.38\linewidth}
                                >{\raggedright\arraybackslash}X@{}}
\toprule
Direction & Representative searches & Authoring agent's assessment \\
\midrule
Tabular and columnar data
  & CSV edge cases; corrupted Parquet metadata
  & Parquet repair appeared \emph{too shallow}. \\
\addlinespace[0.28em]
Database and repository recovery
  & SQLite/WAL salvage; damaged Git objects
  & Git recovery was too close to existing tasks; SQLite/WAL recovery was too tool-specific. \\
\addlinespace[0.28em]
Archive and text recovery
  & Truncated gzip; mixed encodings; malformed NDJSON
  & Format-specific repair risked collapsing into a single utility invocation. \\
\addlinespace[0.28em]
Media and network formats
  & PCAP extraction; EXIF/JPEG repair; MP3/ID3 recovery
  & PCAP was judged \emph{too security-oriented}. \\
\addlinespace[0.28em]
Legacy and heterogeneous data
  & Fixed-width/COBOL records; multi-format ETL
  & Legacy-format ETL required substantial setup without a comparably clear verification target. \\
\addlinespace[0.28em]
\rowcolor{casegreenback}
\textcolor{casegreen}{Binary sensor logs}
  & IoT formats; sync markers; checksums; interrupted writes
  & \textcolor{casegreen}{\textbf{Selected}}. \textcolor{casegreen}{\textbf{Reason}}: grounded, multi-step, not solvable by one command, and deterministically verifiable. \\
\bottomrule
\end{tabularx}

\medskip
\textbf{Search-grounded direction.}
Recover environmental readings from a corrupted binary sensor log using record synchronization and checksum validation.
\end{casesearch}

\casearrow

\begin{caseoutput}{\textcircled{\scriptsize 3}\quad Output $I$ --- Task Instruction}
\small
\raggedright
An EnviroLog-2000 monitoring station stores temperature, humidity, pressure, and CO$_2$ readings in a custom binary format.
A power failure corrupted \path{/app/sensor_data.bin}; the format specification is provided in \path{/app/FORMAT_SPEC.md}.
Recover as many valid readings as possible and write
\path{/app/recovered_readings.csv} with the columns
\texttt{timestamp}, \texttt{temperature}, \texttt{humidity},
\texttt{pressure}, and \texttt{co2}, sorted by timestamp.
Do not trust the header record count, and discard records with invalid checksums or impossible sensor values.
\end{caseoutput}

\casearrow

\begin{caseoutput}{\textcircled{\scriptsize 4}\quad Output $E$ --- Environment and Initial Artifacts}
\small
\renewcommand{\arraystretch}{1.14}
{\small
\begin{tabularx}{\linewidth}{@{}>{\raggedright\arraybackslash}p{0.40\linewidth}X@{}}
\toprule
\path{/app/sensor_data.bin}
  & Corrupted binary log containing intact, truncated, and overwritten records. \\
\path{/app/FORMAT_SPEC.md}
  & Record layout, sync marker, CRC-16 procedure, and physical validity ranges. \\
Dockerfile
  & Ubuntu 24.04 environment with Python and the two task files placed under \path{/app}. \\
\bottomrule
\end{tabularx}
}
\end{caseoutput}

\casearrow

\begin{caseoutput}{\textcircled{\scriptsize 5}\quad Output $V$ --- Verification Scope}
\small
{\small
\begin{tabularx}{\linewidth}{@{}>{\color{casegreen}\bfseries}p{0.055\linewidth}X@{}}
\raisebox{1.65ex}[0pt][0pt]{\(\checkmark\)} & The output CSV exists and has the required schema. \\
\raisebox{1.65ex}[0pt][0pt]{\(\checkmark\)} & Recovered records have the correct count and chronological order. \\
\raisebox{1.65ex}[0pt][0pt]{\(\checkmark\)} & The overwritten interval is detected and excluded. \\
\raisebox{1.65ex}[0pt][0pt]{\(\checkmark\)} & Timestamps and sensor values satisfy the expected ranges and boundary cases. \\
\raisebox{1.65ex}[0pt][0pt]{\(\checkmark\)} & Selected recovered records match their expected values. \\
\end{tabularx}
}
\medskip
\textbf{Implementation:} 11 verifier tests.
\end{caseoutput}

\casearrow

\begin{casecalibration}{\textcircled{\scriptsize 6}\quad Contrastive Calibration Feedback}
\small
\renewcommand{\arraystretch}{1.18}
{\small
\begin{tabularx}{\linewidth}{@{}>{\bfseries}p{0.24\linewidth}p{0.40\linewidth}X@{}}
\toprule
Role & Solver model & Verified outcome \\
\midrule
Stronger solver & DeepSeek-V4-Pro & \textcolor{casegreen}{\textbf{PASS}} \\
Weaker solver & DeepSeek-V4-Flash & \textcolor{casered}{\textbf{FAIL}} \\
\bottomrule
\end{tabularx}
}

\medskip
\noindent\begin{minipage}{\linewidth}
\textbf{Solver self-assessments}\par
\smallskip
{\small
\begin{tabularx}{\linewidth}{@{}>{\bfseries}p{0.20\linewidth}p{0.25\linewidth}>{\raggedright\arraybackslash}X@{}}
\toprule
Role & Difficulty & Condensed assessment \\
\midrule
Stronger
  & Easy--medium
  & The format specification was clear; an inconsistent example CRC required validation against actual records. \\
Weaker
  & Medium
  & Binary-structure parsing and combined CRC/range validation were the main challenges. \\
\bottomrule
\end{tabularx}
}
\end{minipage}
\end{casecalibration}

\section{How Solver Feedback Revises Tasks}
\label{supp:revision-cases}

The following cases show how structured feedback and full interaction trajectories guide task revision during adversarial solver calibration.
Each case presents the initial pass/fail outcomes, the feedback used by the authoring agent, the resulting task revision, and the outcomes after revalidation and re-probing.

\Needspace{6\baselineskip}
\subsection{Removing Procedural Hints after Both Solvers Pass}

\def\casecontinuedtitle{Case B1 (continued)}
\begin{revisioncasebox}
  {Case B1: Recovering Corrupted Transaction Data \hfill Contrastive Calibration}
  {casecontrast}
\small
\textbf{Task.}
Repair corrupted fixed-width transaction records and export a normalized CSV file.

\revisionstage{casegrayback}{Initial Outcome}
\textcolor{casered}{\textbf{Both Pass}} in the first calibration round.
\smallskip

\begin{tabular*}{\linewidth}{@{\extracolsep{\fill}}>{\bfseries}l l c c l@{}}
\toprule
Role & Solver model & Outcome & Steps & Difficulty of self-assessment \\
\midrule
Stronger & DeepSeek-V4-Pro & \textcolor{casegreen}{\textbf{PASS}} & 8 & Easy \\
Weaker & DeepSeek-V4-Flash & \textcolor{casegreen}{\textbf{PASS}} & 17 & Medium \\
\bottomrule
\end{tabular*}

\revisionstage{casegrayback}{Solver Evidence}
The stronger solver rated the task easy because the instruction exposed the exact field positions, corruption patterns, and a worked date-repair example.
The weaker solver identified fixed-width extraction and handling the two disclosed corruption types as the main work.
Their assessments indicated that success depended on implementing an explicit repair procedure rather than diagnosing the malformed records.

\revisionstage{caseorangeback}{Author Revision}
{\small
\begin{tabularx}{\linewidth}{@{}X@{\hspace{0.8em}}X@{}}
\begin{minipage}[t]{\linewidth}
\textbf{BEFORE}\\[0.2em]
\emph{Two types of corruption have been identified:}
extra leading whitespace shifts every field, and invalid dates contain swapped month and day components.
\end{minipage}
&
\begin{minipage}[t]{\linewidth}
\textbf{AFTER REVISION}\\[0.2em]
Corrupted records may have structural issues or invalid field values.
The solver must inspect the data and layout specification, determine what went wrong, and repair the records.
\end{minipage}
\end{tabularx}
}

\revisionstage{casegreenback}{Outcome after Re-probing}
{\centering
\textbf{Stronger solver:} \textcolor{casegreen}{\textbf{PASS}}
\qquad
\textbf{Weaker solver:} \textcolor{casered}{\textbf{FAIL}}
\par}

\begin{center}
\textcolor{casegreen}{\bfseries RETAINED AFTER RE-PROBING}
\end{center}
\end{revisioncasebox}

The initial all-pass pattern exposed an overly explicit solution path rather than merely low task difficulty.
Removing the disclosed repair path preserved the required output while making success depend on diagnosing the corrupted records.

\Needspace{6\baselineskip}
\subsection{Clarifying Comparison Semantics after All Solvers Fail}

\def\casecontinuedtitle{Case B2 (continued)}
\begin{revisioncasebox}
  {Case B2: Comparing Customer Database Exports \hfill Multi-Solver Calibration}
  {casemulti}
\small
\textbf{Task.}
Compare two customer-database exports and report record-level and schema-level differences after normalization.

\revisionstage{casegrayback}{Initial Outcome}
\textcolor{casered}{\textbf{All Fail}} in the first calibration round.
\smallskip

\begin{tabular*}{\linewidth}{@{\extracolsep{\fill}}>{\bfseries}l c c l@{}}
\toprule
Solver model & Outcome & Steps & Difficulty of self-assessment \\
\midrule
GLM-5 & \textcolor{casered}{\textbf{FAIL}} & 50 & Medium--hard \\
Kimi-K2.5 & \textcolor{casered}{\textbf{FAIL}} & 15 & Medium \\
DeepSeek-V4-Flash & \textcolor{casered}{\textbf{FAIL}} & 26 & Medium \\
\bottomrule
\end{tabular*}

\revisionstage{casegrayback}{Solver Evidence}

\textbf{Task.}
The task provides two snapshots of the same customer database:
a reference export and a current export.
The solver must normalize the two files and produce a comparison report that separates schema-level differences from modifications to individual customer records.

\smallskip
{\small
\begin{tabularx}{\linewidth}{@{}>{\bfseries}p{0.17\linewidth}XX@{}}
\toprule
& Reference export & Current export \\
\midrule
Header
  & \path{id,name,email}
  & \path{id,name,email,tier} \\
Record
  & \path{42,Ana,a@x.com}
  & \path{42,Ana,a@x.com,gold} \\
\bottomrule
\end{tabularx}
}

\smallskip
\textbf{Solver output versus verifier expectation.}

{\small
\begin{tabularx}{\linewidth}
{@{}>{\bfseries}p{0.19\linewidth}XX@{}}
\toprule
Report component
  & All three solvers reported
  & Verifier expected \\
\midrule
Schema differences
  & Added column \texttt{tier}.
  & Added column \texttt{tier}. \\
Modified records
  & Customer \texttt{id=42} was modified because the current
    export contains the additional field \texttt{tier=gold}.
  & No modified record because the shared fields
    \texttt{id}, \texttt{name}, and \texttt{email} are unchanged. \\
\bottomrule
\end{tabularx}
}

\smallskip
Because all three solvers produced the same extra record-level
classification, the shared all-fail pattern exposed an underspecified
comparison rule rather than independent implementation errors.

\revisionstage{caseorangeback}{Author Revision}
{\small
\begin{tabularx}{\linewidth}{@{}X@{\hspace{0.8em}}X@{}}
\begin{minipage}[t]{\linewidth}
\textbf{BEFORE}\\[0.2em]
Report schema changes and modified customer records.
Either export may contain columns absent from the other.
\end{minipage}
&
\begin{minipage}[t]{\linewidth}
\textbf{AFTER REVISION}\\[0.2em]
Determine whether a customer record is modified by comparing only
fields shared by both exports.
Treat columns present in only one export as schema changes, not record
modifications.
\end{minipage}
\end{tabularx}
}

\revisionstage{casegreenback}{Outcome after Re-probing}
\textbf{Mixed Outcomes} after re-probing.
\smallskip

{\centering
\textbf{GLM-5:} \textcolor{casegreen}{\textbf{PASS}}
\qquad
\textbf{Kimi-K2.5:} \textcolor{casered}{\textbf{FAIL}}
\qquad
\textbf{DeepSeek-V4-Flash:} \textcolor{casegreen}{\textbf{PASS}}
\par}

\begin{center}
\textcolor{casegreen}{\bfseries RETAINED AFTER RE-PROBING}
\end{center}
\end{revisioncasebox}

The shared failures identified ambiguity in the task instruction rather than a need to simplify the underlying data-processing problem.
The revision resolved the ambiguity without making the task uniformly
easy: two solvers passed after re-probing, while one still failed,
satisfying the multi-solver retention criterion.

\Needspace{6\baselineskip}
\subsection{Generalizing an Overly Prescriptive Verifier after an Inverted Outcome}

\def\casecontinuedtitle{Case B3 (continued)}
\begin{revisioncasebox}
  {Case B3: Securing a Legacy Password Vault \hfill Contrastive Calibration}
  {casecontrast}
\small
\textbf{Task.}
Repair cryptographic vulnerabilities in a command-line password manager while preserving its interface and keeping entries from the legacy vault readable.

\revisionstage{casegrayback}{Initial Outcome}
\textcolor{casered}{\textbf{Inverted Outcome}} in the first calibration round.
\smallskip

\begin{tabular*}{\linewidth}{@{\extracolsep{\fill}}>{\bfseries}l l c c l@{}}
\toprule
Role & Solver model & Outcome & Steps & Difficulty of self-assessment \\
\midrule
Stronger & DeepSeek-V4-Pro & \textcolor{casered}{\textbf{FAIL}} & 40 & Medium \\
Weaker & DeepSeek-V4-Flash & \textcolor{casegreen}{\textbf{PASS}} & 38 & Medium \\
\bottomrule
\end{tabular*}

\revisionstage{casegrayback}{Solver Evidence}
The task required all sensitive fields to be encrypted while keeping existing vault entries readable.
The stronger solver combined the password, notes, and other sensitive fields into one record and encrypted the record as a whole using authenticated encryption.
However, the verifier expected these fields to contain separately encrypted ciphertext and therefore rejected this valid alternative layout.
The weaker solver followed the verifier's expected field-wise layout and passed.
The inverted outcome therefore exposed an overly prescriptive verifier rather than an invalid solution.

\revisionstage{caseorangeback}{Author Revision}
{\small
\begin{tabularx}{\linewidth}{@{}X@{\hspace{0.8em}}X@{}}
\begin{minipage}[t]{\linewidth}
\textbf{BEFORE}\\[0.2em]
The verifier expected named fields to contain separately encoded
\texttt{nonce:ciphertext:tag} values and swapped the
\texttt{password} fields to test context binding.
\end{minipage}
&
\begin{minipage}[t]{\linewidth}
\textbf{AFTER REVISION}\\[0.2em]
The verifier discovers the encrypted payload independently of its
field name and tests authenticated encryption, context binding, and
legacy migration without prescribing the vault's object layout.
\end{minipage}
\end{tabularx}
}

\revisionstage{casegreenback}{Outcome after Re-probing}
{\centering
\textbf{Stronger solver:} \textcolor{casegreen}{\textbf{PASS}}
\qquad
\textbf{Weaker solver:} \textcolor{casered}{\textbf{FAIL}}
\par}

\begin{center}
\textcolor{casegreen}{\bfseries RETAINED AFTER RE-PROBING}
\end{center}
\end{revisioncasebox}

The inverted outcome exposed a brittle verifier rather than an invalid solution.
Generalizing the verifier preserved the required security properties while allowing alternative valid storage layouts.
After the verifier was revised, a subsequent calibration round produced a stronger-solver pass and a weaker-solver failure, satisfying the contrastive retention criterion.

\paragraph{Summary.}
Verified pass/fail outcomes determine whether a candidate satisfies its retention criterion, but do not explain why the observed outcome pattern occurs.
Across these cases, feedback summaries and full interaction trajectories reveal problems such as overly explicit instructions, underspecified requirements, and overly prescriptive verifiers, thereby guiding task revision.

\Needspace{10\baselineskip}
\section{\scaffold{}}
\label{supp:calibforge-eval}

\Needspace{6\baselineskip}
\subsection{Tool Interface}

\scaffold{} is a minimal code-agent scaffold implemented using AweAgent~\citep{aweagent2026} for trajectory distillation and Terminal-Bench evaluation.
Following the DeepSeek-V4 code-agent evaluation setting~\citep{xu2026deepseek}, it exposes only bash, file-editing, and finish tools under a compact prompt.

\begin{table}[H]
    \centering
    \small
    \setlength{\tabcolsep}{4pt}
    \renewcommand{\arraystretch}{1.12}
    \begin{tabularx}{\linewidth}{@{}>{\raggedright\arraybackslash\bfseries}p{0.21\linewidth}
                                    >{\raggedright\arraybackslash}p{0.37\linewidth}
                                    >{\raggedright\arraybackslash}X@{}}
        \toprule
        \textbf{Tool} & \textbf{Parameters} & \textbf{Function} \\
        \midrule
        \cellcolor{tableblue}\texttt{execute\_bash}
          & \texttt{command}: string (required);
            \texttt{timeout}: number (optional)
          & Executes one Bash command in the persistent task runtime and returns its output and execution status. \\
          \addlinespace[0.5em]
        \cellcolor{tableblue}\texttt{str\_replace\_editor}
          & \texttt{command}: \texttt{view}, \texttt{create}, \texttt{str\_replace}, or \texttt{insert};
            \texttt{path}: string (both required);
            operation-specific \texttt{file\_text}, \texttt{old\_str}, \texttt{new\_str}, \texttt{insert\_line}, or \texttt{view\_range}
          & Views, creates, and edits files in the persistent task runtime. \\
          \addlinespace[0.5em]
        \cellcolor{tableblue}\texttt{finish}
          & No arguments
          & Ends the interaction; task success is subsequently determined by running the verifier on the final runtime state. \\
        \bottomrule
    \end{tabularx}
    \caption{Function-tool interface exposed by \scaffold{}.
    Operation-specific parameters are used only by the corresponding \texttt{str\_replace\_editor} operation.}
    \label{tab:supp-eval-tools}
\end{table}

\Needspace{6\baselineskip}
\subsection{Prompt Templates}

\begin{promptbox}[colback=caseblueback,colframe=caseblue,colbacktitle=caseblue,title after break={System Prompt (continued)}]{CalibForge-Eval System Prompt}
You are a capable code and command-line agent operating in a Linux runtime environment. Complete the user's technical task by inspecting the environment, editing files, running commands, and verifying the result. Use the available tools directly whenever an action is needed.

<WORKFLOW>
1. Read the task carefully and identify the concrete deliverable, required output format, exact paths, field names, variable names, version constraints, and any named interfaces, scripts, files, or services. Treat these details as hard requirements.
2. Survey the environment before changing it. Check the working directory and nearby resources, but do not assume all relevant files are under the initial directory.
3. Prefer existing project workflows and task-provided interfaces over building local substitutes. If a CLI, script, API, or data format is mentioned, inspect its help, documentation, source, or sample data before using it.
4. Pay attention to implicit constraints in the task wording, especially constraints about exact outputs, allowed files, dependency versions, or how simple the requested solution should be.
5. Make focused changes that solve the task. Modify intended files directly instead of creating alternate copies with suffixes such as _fixed, _new, or _test.
6. Keep deliverable directories clean. Temporary scripts, downloaded files, build outputs, compiled binaries, caches, and exploratory artifacts should stay outside the final deliverable location unless the task explicitly asks for them.
7. Verify the result with task-relevant commands whenever possible. If full verification is expensive or unavailable, run targeted smoke checks that exercise the requested behavior.
8. Finish only after the environment is in the final state needed for grading.
</WORKFLOW>

<TOOL_USE>
Use the bash tool for shell commands, builds, package installation, data processing, service checks, and running verification commands. Use the file editor tool for precise text-file inspection and edits. Prefer absolute paths for file operations.

For text edits, view enough context first and ensure replacements are exact and unique. If several edits to one file are needed, batch them when practical. For multi-line code or structured text, prefer the file editor tool over fragile shell quoting; if you generate files with shell commands, read them back before running them.

Do not add explanatory documentation or extra artifacts unless the task asks for them or they are required by the solution.
</TOOL_USE>

<LONG_RUNNING_WORK>
Some tasks require slow downloads, dependency installation, compilation, model loading, or large data processing. For commands that may take longer than the default command timeout, pass an appropriate timeout value to the bash tool. If a long command times out or produces partial logs, inspect the logs, adjust the approach, and retry with a clearer command, a longer timeout, or a background process with log polling.

When dependencies are missing, first look for existing project files such as requirements.txt, pyproject.toml, package.json, Cargo.toml, Makefile, or README instructions. Prefer the repository's declared setup path over ad-hoc package installs, but install missing dependencies when necessary to complete the task.

Network downloads may be slow or partially blocked. For GitHub, Hugging Face, PyPI, apt, video sites, and large artifacts, first identify whether the issue is speed, availability, permission, rate limiting, or anti-bot behavior. For slow downloads, try reliable mirrors, resumable commands, or smaller range checks, then verify the downloaded file's size and format. For video-site or anti-bot failures, look for legitimate mirrors, archives, or alternate sources before relying on an approximation.
</LONG_RUNNING_WORK>

<VERIFICATION>
Do not mark the task complete just because a file was created, a command started, or an import succeeded once. Check the resulting files, imports, command output, tests, service behavior, or structured data that matter for the task. For JSON, CSV, databases, archives, compressed files, or generated reports, load or preview the artifact with an appropriate tool and confirm the schema and contents match the requirement.

If a check fails, use the failure message to continue debugging instead of finishing. If the same approach fails repeatedly, inspect the relevant files, logs, or help output and switch to a different method rather than retrying blindly.

When running provided checks or tests, confirm that the command you ran is the intended entry point. Empty output or a zero exit code is not enough if the task requires a specific file, exact one-line answer, schema, version, or absence of extra files.
</VERIFICATION>

<SAFETY_AND_SCOPE>
Work only on changes needed for the task. Avoid destructive operations unless they are clearly required. Do not rely on hidden grader behavior or hard-code answers for a specific verifier; solve the stated task in the environment. The user will not provide interactive clarification during the run, so make reasonable assumptions and keep working toward a verified final state.
</SAFETY_AND_SCOPE>

When the task is complete, call the finish tool. Do not finish merely by saying the task is done in text.
\end{promptbox}

\begin{promptbox}[unbreakable,colback=caseblueback,colframe=caseblue,colbacktitle=caseblue]{CalibForge-Eval User Prompt Template}
Task:
{instruction}

Working directory: {workdir}

Complete the task in the current runtime environment. The final state of this environment will be evaluated after you call finish.
\end{promptbox}

\section{Benchmark Decontamination}
\label{supp:decontamination}

Before trajectory distillation and model training, we compare every \papername{} candidate task against the evaluation instances in Terminal-Bench 2.1, SWE-bench Pro, and Doc2Repo~\citep{merrill2026terminal,deng2025swe,chen2026beyondswe}.
We first remove candidates whose instructions contain an exact 14-gram match with an evaluation instruction.
We then compute 5-shingle Jaccard similarity over task instructions and available verifier or test code.
Before computing shingles, we lowercase and tokenize the text, remove benchmark-specific boilerplate, and replace URLs, absolute file paths, and numeric values with canonical placeholders.
The core similarity thresholds are \(0.30\) for task instructions and \(0.45\) for verifier or test code.

Because contamination may remain despite surface-level differences in task instructions, we combine these similarity signals with structural evidence, including shared output paths, overlapping test functions, and high-risk task-family matches.
Any candidate flagged against at least one of the three evaluation benchmarks is removed before trajectory distillation and model training.

\section{Training Details}
\label{supp:training-details}

\Needspace{6\baselineskip}
\subsection{Supervised Fine-Tuning}
\label{supp:sft-details}

We train Qwen3-30B-A3B-Instruct and Qwen3.5-35B-A3B~\citep{qwen3technicalreport,qwen3.5} using full-parameter, multi-turn SFT implemented with LLaMA-Factory~\citep{zheng2024llamafactory}.
Both models use the shared configuration in Table~\ref{tab:supp-training-details}, and we report the final checkpoint after 10 epochs.

The multi-solver ablation in Table~\ref{tab:solver_ablation} uses the original 1,300-task collection.
A subsequent audit identified 37 tasks that had been incorrectly retained despite not satisfying the required mixed pass/fail criterion.
These tasks were removed before training the final models, leaving 1,263 multi-solver tasks in the final 5,431-task collection.

\begin{table}[H]
\centering
\small
\setlength{\tabcolsep}{5pt}
\renewcommand{\arraystretch}{1.03}
\begin{tabularx}{\linewidth}{@{}>{\raggedright\arraybackslash}p{0.22\linewidth}>{\raggedright\arraybackslash}p{0.20\linewidth}>{\raggedright\arraybackslash}p{0.27\linewidth}>{\raggedright\arraybackslash}X@{}}
\toprule
\rowcolor{tableblue}\textbf{Hyperparameter} & \textbf{Value} & \textbf{Hyperparameter} & \textbf{Value} \\
\midrule
Optimizer & AdamW (\(\beta_1{=}0.9\), \(\beta_2{=}0.999\)) & Per-device train batch size & 1 \\
Learning rate & \(1.0\times10^{-5}\) & Gradient accumulation steps & 4 \\
LR scheduler & cosine & Global batch size & 128 \\
Warmup ratio & 0.05 & Context length & 131{,}072 \\
Weight decay & 0.0 & Precision & bf16 \\
Maximum gradient norm & 1.0 & GPUs & 64 \(\times\) NVIDIA H20 \\
Tensor parallel size & 2 & Epochs & 10 \\
\bottomrule
\end{tabularx}
\caption{Supervised fine-tuning hyperparameters and hardware used for both \papername{} student models.}
    \label{tab:supp-training-details}
\end{table}

\Needspace{6\baselineskip}
\subsection{Reinforcement Learning}
\label{supp:rl-details}
\begin{figure}[H]
\centering
\includegraphics[width=0.76\linewidth]{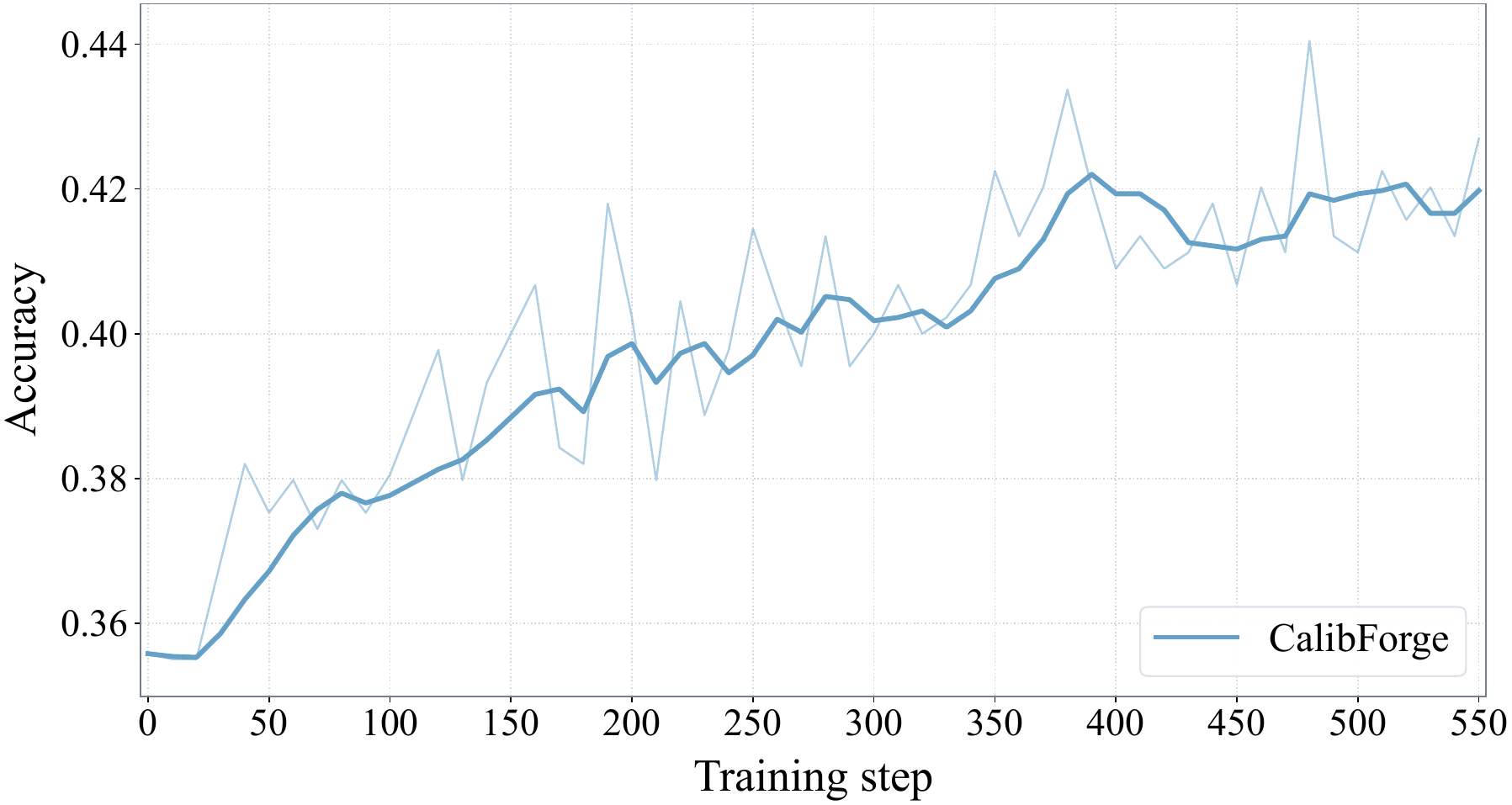}
\caption{Extended GRPO training curve for \papername{} with Qwen3.5-35B-A3B on Terminal-Bench 2.1.
The thin line shows checkpoint accuracy averaged over five evaluation runs; the thick line shows a trailing five-checkpoint mean.}
\label{fig:grpo_calibforge_extended}
\end{figure}

We train Qwen3.5-35B-A3B with fully asynchronous GRPO using slime~\citep{slime_github}, with AweAgent~\citep{aweagent2026} generating terminal-task rollouts.
The \papername{}, No Solver, and Single Solver runs each use 1,024 tasks and share the training hyperparameters in Table~\ref{tab:supp-rl-details}.
Verified binary outcomes are mapped to rewards of $+1$ for success and $-1$ for failure; advantages are normalized within each group of 16 rollouts.
Figure~\ref{fig:grpo_calibforge_extended} shows the \papername{} training curve through 550 steps.

\begin{table}[H]
\centering
\small
\setlength{\tabcolsep}{5pt}
\renewcommand{\arraystretch}{1.03}
\begin{tabularx}{\linewidth}{@{}>{\raggedright\arraybackslash}p{0.22\linewidth}>{\raggedright\arraybackslash}p{0.20\linewidth}>{\raggedright\arraybackslash}p{0.27\linewidth}>{\raggedright\arraybackslash}X@{}}
\toprule
\rowcolor{tableblue}\textbf{Hyperparameter} & \textbf{Value} & \textbf{Hyperparameter} & \textbf{Value} \\
\midrule
Optimizer & Adam & Tasks per rollout batch & 16 \\
Learning rate & $1.0\times10^{-6}$ & Rollouts per task & 16 \\
LR scheduler & constant & Global batch size & 256 rollouts \\
Adam $(\beta_1,\beta_2)$ & $(0.9,0.98)$ & Sampling temperature & 1.0 \\
Weight decay & 0.1 & Top-$p$ & 0.95 \\
PPO clip $(\epsilon_{\mathrm{low}},\epsilon_{\mathrm{high}})$ & $(0.2,0.28)$ & Top-$k$ & disabled \\
KL penalty coefficient & 0.0 & Context limit & 128,000 tokens \\
Entropy coefficient & 0.0 & Rollout response budget & 128,000 tokens \\
Importance-weight clipping & $[0,2]$ & Per-call generation limit & 32,768 tokens \\
Group standard-deviation normalization & enabled & Agent step limit & 500 \\
Global advantage whitening & disabled & Per-task timeout & 3,600 s \\
Zero-variance group filtering & disabled & Rollout seed & 42 \\
\bottomrule
\end{tabularx}
\caption{Shared GRPO hyperparameters for the three 1,024-task training subsets. Truncated importance sampling corrects for differences between the rollout and training policies.}
\label{tab:supp-rl-details}
\end{table}

\begingroup
\renewtcolorbox{revisioncasebox}[2]{
  enhanced, breakable, colback=white, colframe=#2,
  colbacktitle=#2, coltitle=white, title={#1}, fonttitle=\bfseries\raggedright, title after break={\casecontinuedtitle},
  boxrule=0.8pt, arc=1mm, left=4mm, right=4mm, top=2.5mm,
  bottom=2.5mm, before skip=6pt, after skip=5pt,
  before upper={\setlength{\parskip}{0pt}}
}
\renewcommand{\revisionstage}[2]{%
  \par\addvspace{3pt}%
  \noindent\colorbox{#1}{%
    \parbox{\dimexpr\linewidth-2\fboxsep\relax}{\textbf{#2}}%
  }%
  \par\nobreak\vspace{1pt}%
}
\section{Failure Analysis of Trained Models}
\label{supp:failure-analysis}

For qualitative failure analysis, we inspect trajectories from \textbf{three earlier Terminal-Bench 2.0 evaluation runs} of both \papername{}-30B-A3B and \papername{}-35B-A3B under the same \scaffold{} setting.
The cases below were selected because their decisive actions, final artifacts, and verifier results establish a clear failure mechanism.
Together, these cases expose distinct failure mechanisms that are not visible from aggregate benchmark scores alone.

\Needspace{6\baselineskip}
\subsection{Reasoning without Producing the Required Artifact}

\def\casecontinuedtitle{Case F1 (continued)}
\begin{revisioncasebox}
  {Case F1: \texttt{regex-log}}{caseblue}
\small
\textbf{Task Requirement.}
Construct a regular expression that returns the last valid date from each log line containing a valid IPv4 address, and save it to \path{/app/regex.txt} for evaluation with Python \texttt{re.findall}.

\revisionstage{casegrayback}{Observed Outcomes}
{\small
\begin{tabular*}{\linewidth}{@{\extracolsep{\fill}}lccc@{}}
\toprule
Model & Run 1 & Run 2 & Run 3 \\
\midrule
\papername{}-30B-A3B
  & \textcolor{casered}{\textbf{FAIL}}
  & \textcolor{casered}{\textbf{FAIL}}
  & \textcolor{casered}{\textbf{FAIL}} \\
\papername{}-35B-A3B
  & \textcolor{casegreen}{\textbf{PASS}}
  & \textcolor{casegreen}{\textbf{PASS}}
  & \textcolor{casegreen}{\textbf{PASS}} \\
\bottomrule
\end{tabular*}
}

\revisionstage{casegrayback}{Trajectory Evidence}
The 30B runs repeatedly constructed candidate expressions and exercised them in scratch scripts, but none wrote the required \path{/app/regex.txt} before reaching the time limit.
In contrast, each 35B run materialized the selected expression in the required file and tested it using the same \texttt{re.findall} interface specified by the task, including lines with multiple dates and invalid IPv4-like strings.

\par\smallskip
\noindent\begin{minipage}{\linewidth}
\revisionstage{casegrayback}{Verifier Evidence}
All three 30B runs failed at the first check because \path{/app/regex.txt} did not exist.
All three 35B runs passed the verifier.
\end{minipage}

\revisionstage{caseorangeback}{Diagnosis}
The failed runs made substantial progress on the regex itself but did not convert that progress into the required deliverable.
This case separates solution exploration from task completion: useful intermediate reasoning is insufficient unless the agent commits the final artifact and checks it through the specified interface.
\end{revisioncasebox}

\Needspace{6\baselineskip}
\subsection{Committing to Partial Forensic Evidence}

\def\casecontinuedtitle{Case F2 (continued)}
\begin{revisioncasebox}
  {Case F2: \texttt{password-recovery}}{caseblue}
\small
\textbf{Task Requirement.}
Recover a deleted 23-character password that starts with \texttt{8XD} and ends with \texttt{W54}, and write each candidate password to \path{/app/recovered_passwords.txt}.

\revisionstage{casegrayback}{Observed Outcomes}
{\small
\begin{tabular*}{\linewidth}{@{\extracolsep{\fill}}lccc@{}}
\toprule
Model & Run 1 & Run 2 & Run 3 \\
\midrule
\papername{}-30B-A3B
  & \textcolor{casered}{\textbf{FAIL}}
  & \textcolor{casered}{\textbf{FAIL}}
  & \textcolor{casered}{\textbf{FAIL}} \\
\papername{}-35B-A3B
  & \textcolor{casegreen}{\textbf{PASS}}
  & \textcolor{casegreen}{\textbf{PASS}}
  & \textcolor{casegreen}{\textbf{PASS}} \\
\bottomrule
\end{tabular*}
}

\revisionstage{casegrayback}{Trajectory Evidence}
Two 30B runs wrote \texttt{PASSWORD=8XDK7VB3BV4W54}, treating the stated length as if it included the \texttt{PASSWORD=} prefix; the third run produced no output file before timing out.
The successful 35B runs instead combined the recoverable prefix \texttt{8XDP5Q2RT9Z} with the separate suffix fragment \texttt{K7VB3BV4WW54}, then explicitly checked the reconstructed password's length, prefix, suffix, and character set.

\revisionstage{casegrayback}{Verifier Evidence}
The two completed 30B artifacts did not contain the recovered password, and the remaining run lacked the required file.
All three 35B artifacts contained the same valid 23-character password and passed the verifier.

\revisionstage{caseorangeback}{Diagnosis}
The failed runs committed to a locally plausible fragment after misreading a global length constraint.
Successful recovery required integrating evidence from separate regions of the forensic image and rechecking the combined result against every stated constraint.
\end{revisioncasebox}

\Needspace{6\baselineskip}
\subsection{Changing State before Preserving Recovery Evidence}

\def\casecontinuedtitle{Case F3 (continued)}
\begin{revisioncasebox}
  {Case F3: \texttt{db-wal-recovery}}{caseblue}
\small
\textbf{Task Requirement.}
Repair an encrypted SQLite write-ahead log (WAL), recover all 11 records rather than the five records in the base database, and write them to \path{/app/recovered.json}.

\revisionstage{casegrayback}{Observed Outcomes}
{\small
\begin{tabular*}{\linewidth}{@{\extracolsep{\fill}}lccc@{}}
\toprule
Model & Run 1 & Run 2 & Run 3 \\
\midrule
\papername{}-30B-A3B
  & \textcolor{casered}{\textbf{FAIL}}
  & \textcolor{casered}{\textbf{FAIL}}
  & \textcolor{casered}{\textbf{FAIL}} \\
\papername{}-35B-A3B
  & \textcolor{casered}{\textbf{FAIL}}
  & \textcolor{casered}{\textbf{FAIL}}
  & \textcolor{casered}{\textbf{FAIL}} \\
\bottomrule
\end{tabular*}
}

\revisionstage{casegrayback}{Trajectory Evidence}
Across all six runs, the agent opened \texttt{main.db} with SQLite before copying or decrypting \texttt{main.db-wal}.
SQLite returned the five base records and removed the unreadable WAL from the working directory.
Subsequent commands could therefore no longer inspect or repair the file containing the six additional records.
The 30B runs timed out without a valid recovery artifact, while the 35B runs attempted to reconstruct 11 records from the remaining base database.

\revisionstage{casegrayback}{Verifier Evidence}
The 30B runs failed because the required recovery output was absent.
The 35B outputs passed basic JSON and schema checks but failed both the record-completeness check and the check that the WAL had actually been decrypted.

\revisionstage{caseorangeback}{Diagnosis}
The first database query was not a read-only observation: it changed the state needed for the recovery.
This case exposes the importance of preserving volatile evidence before using an application that may checkpoint, delete, or otherwise rewrite sidecar files.
\end{revisioncasebox}

\paragraph{Summary.}
Together, these cases reveal three distinct failure modes: failing to produce the required artifact, committing prematurely to partial evidence, and modifying mutable state before preserving recovery evidence.
Trajectory inspection complements aggregate scores by showing how plausible intermediate progress can still fail to produce a verifier-confirmed solution.

\endgroup
\FloatBarrier

\section{Historical Terminal-Bench 2.0 Results}
\label{supp:tb20-results}
\begingroup
\setlength{\intextsep}{6pt}

\begin{table}[H]
\centering
\small
\setlength{\tabcolsep}{10pt}
\begin{tabular}{@{}lcc@{}}
\toprule
\textbf{Training Data} & \textbf{Qwen3-30B-A3B-Instruct} & \textbf{Qwen3.5-35B-A3B} \\
\midrule
Base Model & \msem{7.87}{0.00} & \msem{38.58}{1.87} \\
Endless Terminals & \msem{19.48}{3.00} & -- \\
CLI-Gym & \msem{23.22}{2.70} & -- \\
SETA-Env & \msem{23.22}{0.75} & -- \\
TermiGen & \msem{23.60}{1.12} & \msem{40.07}{0.99} \\
TerminalTraj & \underline{\msem{26.22}{0.75}} & \underline{\msem{40.82}{0.75}} \\
\textbf{\papername{}} & \textbf{\msem{32.58}{1.12}} & \textbf{\msem{47.57}{0.99}} \\
\bottomrule
\end{tabular}
\caption{Additional results on Terminal-Bench 2.0.}
\label{tab:tb20_results}
\end{table}

\begin{figure}[H]
\centering
\begin{subfigure}[t]{0.48\textwidth}
\centering
\includegraphics[width=0.90\linewidth]{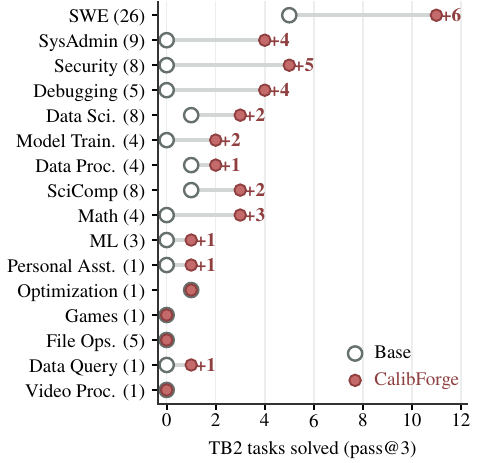}
\caption{\papername{}-30B-A3B}
\end{subfigure}\hfill
\begin{subfigure}[t]{0.48\textwidth}
\centering
\includegraphics[width=0.90\linewidth]{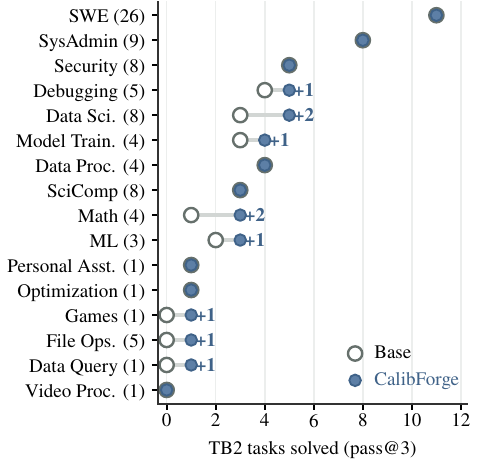}
\caption{\papername{}-35B-A3B}
\end{subfigure}
\caption{Per-category results on Terminal-Bench 2.0.}
\label{fig:per_category_tb20}
\end{figure}
\endgroup
\FloatBarrier

\end{document}